\documentclass[acmtog]{acmart}

\usepackage{booktabs} % For formal tables
\acmPrice{15.00}

\usepackage{multirow}

\usepackage{etoolbox}
\makeatletter
\patchcmd\@combinedblfloats{\box\@outputbox}{\unvbox\@outputbox}{}{%
	\errmessage{\noexpand\@combinedblfloats could not be patched}%
}%\emph{}
\makeatother

\setcopyright{acmcopyright}
\acmJournal{TOG}
\acmYear{2018}
\acmVolume{37}
\acmNumber{6}
\acmMonth{11}
\acmArticle{259}
\acmDOI{10.1145/3272127.3275081}

\setcitestyle{square}

\begin{document}
% Title portion
\title{Image Smoothing via Unsupervised Learning}

% Authors.
\author{Qingnan Fan}
\authornote{{Part of this work was done during Qingnan Fan's internship at MSRA.}}
\affiliation{
  \institution{Shandong University, Beijing Film Academy}}
\email{fqnchina@gmail.com}

\author{Jiaolong Yang}
\affiliation{%
  \institution{Microsoft Research Asia}}
\email{jiaoyan@microsoft.com}

\author{David Wipf}
\affiliation{%
  \institution{Microsoft Research Asia}}
\email{davidwip@microsoft.com}

\author{Baoquan Chen}
\affiliation{%
  \institution{Peking University, Shandong University}}
\email{baoquan@pku.edu.cn}

\author{Xin Tong}
\affiliation{%
  \institution{Microsoft Research Asia}}
\email{xtong@microsoft.com}

% This command defines the author string for running heads.
\renewcommand{\shortauthors}{Qingnan, F. et al}

\graphicspath{{./images/teaser/}{./images/comp_lp_difference/}{./images/comp_state_of_the_art/}{./images/comp_self/comp_optimization/}
{./images/comp_self/comp_edge/}{./images/comp_self/comp_flatten/}{./images/demo/detail_enhancement/}{./images/demo/detail_enhancement_enhance/}
{./images/demo/stylization/}{./images/demo/texture_smoothing/}{./images/demo/background_blur/}{./images/comp_self/parameter/}{./images/comp_self/woResidual/}
{./images/optimization/}{./images/failure_case/}}

\begin{teaserfigure}
\vspace{-1mm}
\begin{center}
\setlength{\tabcolsep}{1pt}
\begin{tabular}{cccccccc}
\includegraphics[height=3.3cm,trim={2cm 0 0 0},clip]{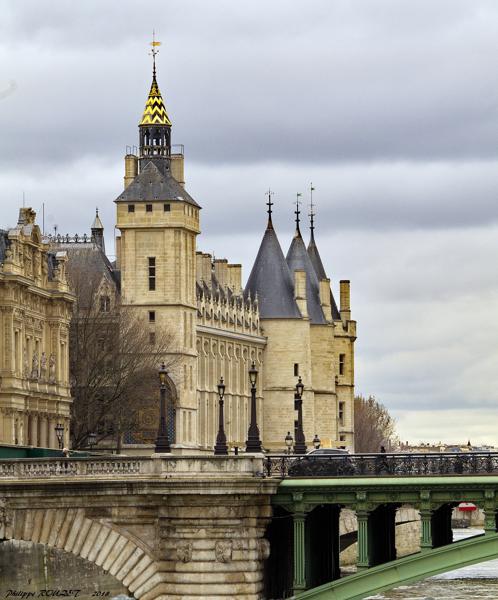}
&\includegraphics[height=3.3cm,trim={2cm 0 0 0},clip]{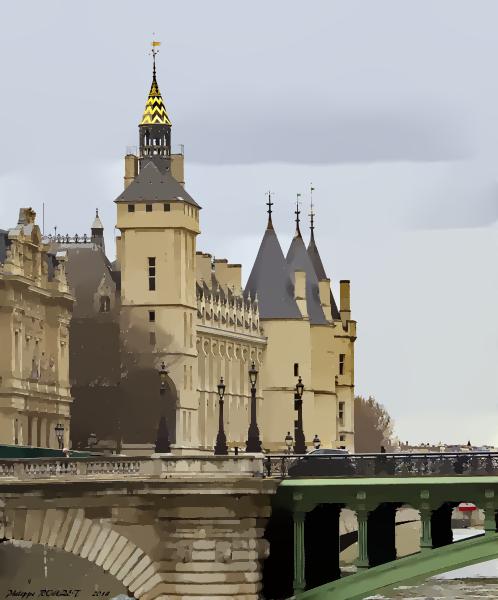}
&\includegraphics[height=3.3cm,trim={2cm 0 0 0},clip]{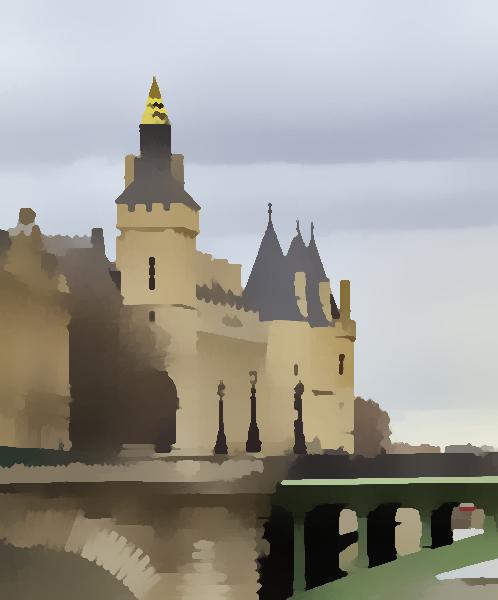}
&\includegraphics[height=3.3cm,trim={2cm 0 0 0},clip]{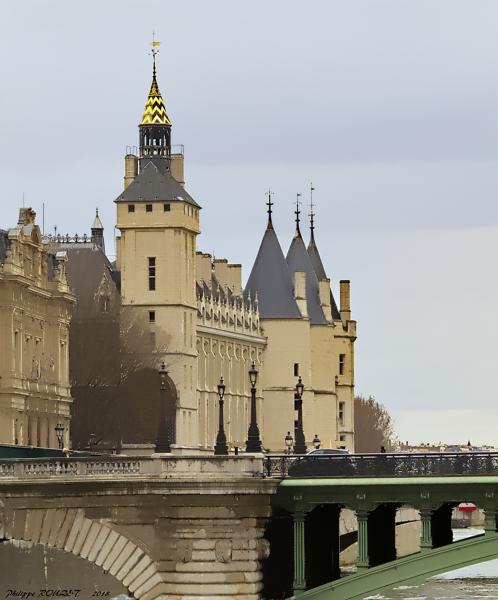}
&\includegraphics[height=3.3cm,trim={0 0 0 0},clip]{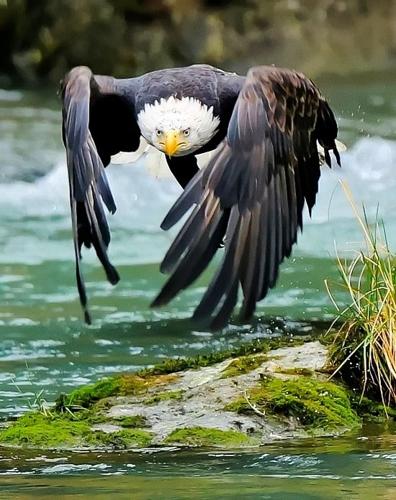}
&\includegraphics[height=3.3cm,trim={0 0 0 0},clip]{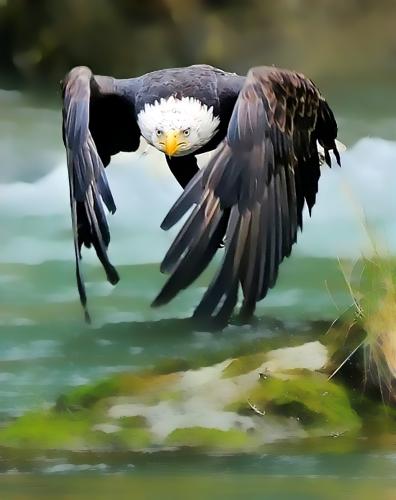}
&\includegraphics[height=3.3cm,trim={0 0 0 0},clip]{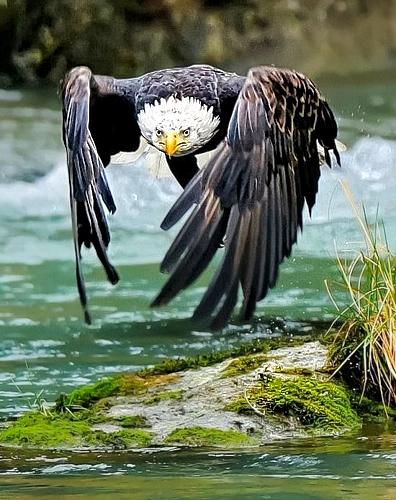}
\\
Input & $L_0$ & SGF & Ours & Input & BG smooth & FG enhance
%\small{Input} & \small{$L_0$} & \small{SGF} & \small{Ours} & \small{Input} & \small{Background smooth} & \small{Foreground enhance}
\\
\end{tabular}
\end{center}
\vspace{-3mm}
\caption{Two applications of traditional edge-preserving image smoothing. \emph{Left 4 panels}: Elimination of low-amplitude details while maintaining high-contrast edges using our method and representative traditional methods $L_0$ \cite{L0smooth} and SGF \cite{zhang2015segment}.  $L_0$ regularization has a strong flattening effect. However, the side effect is that some spurious edges arise in local regions with smooth gradations, such as those on the cloud.  SGF is dedicated to elimination of fine-scale high-contrast details while preserving large-scale salient structures. However, semantically-meaningful information such as the architecture and flagpole can be over-smoothed. In contrast, our result exhibits a more appropriate, targeted balance between color flattening and salient edge preservation.  \emph{Right 3 panels}: Content-aware image manipulation.  Using minimal modification of the guidance image in our proposed pipeline, we are able to implement background (BG) smoothing or foreground (FG) enhancement/exaggeration via a single deep network.  More smoothing effects and applications can be found in Section \ref{sec:applications}.}
\label{figure:teaser}
\vspace{3mm}
\end{teaserfigure}

%It presents one unified image smoothing framework for various applications, which is implemented via optimizing  one flexible objective function through a deep neural network in an unsupervised, label-free setting.

\begin{abstract}
Image smoothing represents a fundamental component of many disparate computer vision and graphics applications. In this paper, we present a unified unsupervised (label-free) learning framework that facilitates generating flexible and high-quality smoothing effects by directly learning from data using deep convolutional neural networks (CNNs). The heart of the design is the training signal as a novel energy function that includes an edge-preserving regularizer which helps maintain important yet potentially vulnerable image structures, and a spatially-adaptive $L_p$ flattening criterion which imposes different forms of regularization onto different image regions for better smoothing quality. We implement a diverse set of image smoothing solutions employing the unified framework targeting various applications such as, image abstraction, pencil sketching, detail enhancement, texture removal and content-aware image manipulation, and obtain results comparable with or better than previous methods. Moreover, our method is extremely fast with a modern GPU (e.g, 200 fps for 1280$\times$720 images). Our codes and model are released in \url{https://github.com/fqnchina/ImageSmoothing}.
\end{abstract}

%keywords

\maketitle

\section{Introduction}

The goal of image smoothing is to eliminate unimportant fine-scale details while maintaining primary image structures. This technique has a wide range of applications in computer vision and graphics, such as tone mapping, detail enhancement, and image abstraction.

Image smoothing has been extensively studied in the past. The early literature was dominated by filtering-based approaches \cite{anisotropic,BLF98,weiss2006fast,paris2006fast,fattal2009edge,DT,chen2007real} due to their simplicity and efficiency. In recent years, smoothing algorithms based on global optimization have gained much popularity due to their superior smoothing results~\cite{WLS,FGS,L0smooth,L1smooth,RTV,liu2017semi}.
Despite the great improvements, however, their smoothing results are still not perfect, and no existing algorithm can serve as an image smoothing panacea for various applications. Moreover, these approaches are often very time-consuming. With the increasing power of modern GPUs and the enormous growth of deep convolutional neural networks (CNNs), there is an emerging interest in employing CNNs as surrogate smoothers in lieu of the costly optimization-based approaches~\cite{xu2015,liu2016,fan2017generic,chen2017fast}. These methods train CNNs in a fully supervised manner where the target outputs are generated by existing smoothing methods.
While substantial speed-up can be achieved, they still produce (approximations of) extant smoothing effects.

In this work, we seek to generate flexible and superior smoothing effects by directly learning from data. We leverage a CNN to do so, such that our method not only features the learned smoothing effects that are more appealing, but also enjoys a fast speed. However, the desired smoothing results (ground-truth labels) for supervising the training are difficult to obtain. Dense manually labeling for a large volume of training images is costly and cumbersome. To circumvent this issue, we design the training signal as an energy function, similar to the optimization-based methods, and train our method in an \emph{unsupervised}, label-free setting.

We carefully designed our energy function to achieve quality smoothing effects in a unified unsupervised-learning framework.
First, to explicitly fortify important image structures that may be weakened by the flattening operator, we include a criterion that minimizes the masked quadratic difference between two guidance maps computed from the  input image and the smoothed estimate respectively. The guidance maps are formulated as edge responses, and the masks are computed using simple edge detection heuristics and can be manually modified further if desired. Second, we identified that many previous methods apply a fixed $L_p$-norm flattening criterion across the entire image, which may be detrimental to the smoothing quality. We instead introduce a spatially-adaptive $L_p$ flattening criterion whereby the specific value of $p$ is varied across images in accordance with the guidance maps. Given that $p=2$ tends to smooth out edges while $p \in (0,1]$ largely preserves them, the guidance maps allow different image regions to receive different regularizations most appropriate for handling local structural conditions. Importantly, we can apply application-specific guidance maps which allow for the seamless implementation of multiple different flattening effects.

We test our method on edge-preserving smoothing and various applications including image abstraction, pencil sketching, detail magnification, texture removal and content-ware image manipulation to show its effectiveness and versatility. Broadly speaking, the contribution of this paper can be distilled as follows:
\begin{itemize}
	\item We introduce an unsupervised learning framework for image smoothing. Unlike previous methods, we do not need ground-truth labels for training and we can jointly learn from any sufficiently diverse corpus of images to achieve the desired performance.
	\item We are able to implement multiple different image smoothing solutions in a single framework, and obtain results comparable with or better than previous methods. A novel image smoothing objective function is proposed to achieve this which is built upon a spatially-adaptive $L_p$ flattening criterion and an edge-preserving regularizer.
	\item Our new method is based on a convolutional neural network and its computational footprint is far below most previous methods. For example, processing a 1280$\times$720 image takes only 5ms on a modern GPU.
\end{itemize}

\setlength{\tabcolsep}{1pt}
\begin{figure*}[htp]
\begin{center}
\begin{tabular}{cccccc}

\includegraphics[width=2.9cm,trim={0 0 0 0},clip]{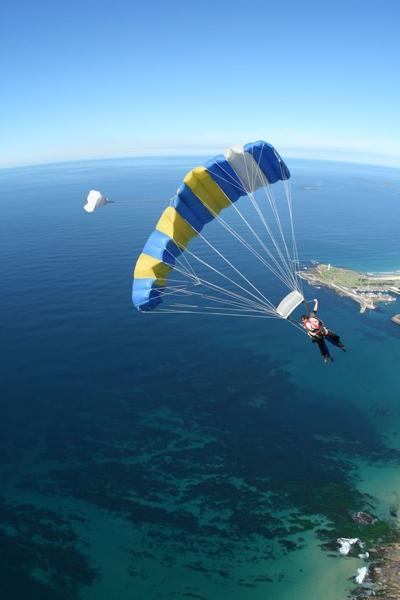}
&\includegraphics[width=2.9cm,trim={0 0 0 0},clip]{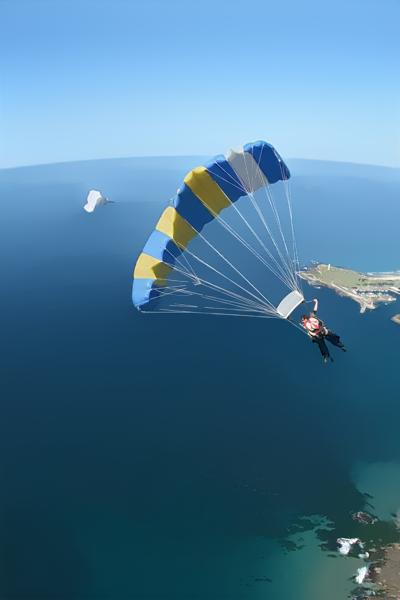}
&\includegraphics[width=2.9cm,trim={0 0 0 0},clip]{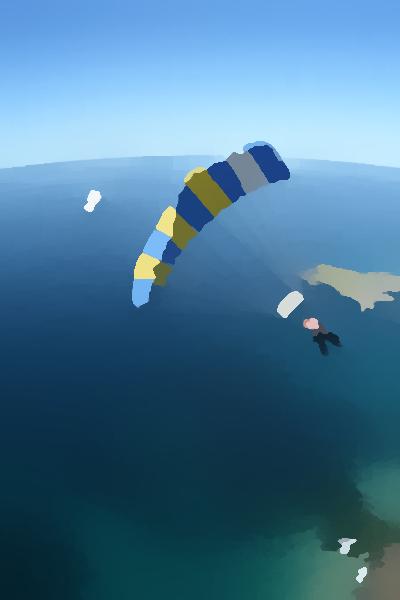}
&\includegraphics[width=2.9cm,trim={0 0 0 0},clip]{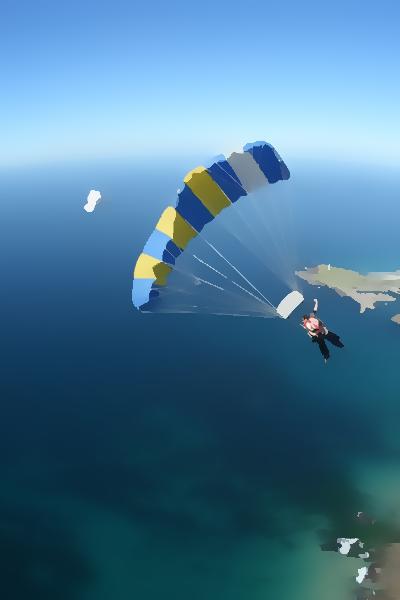}
&\includegraphics[width=2.9cm,trim={0 0 0 0},clip]{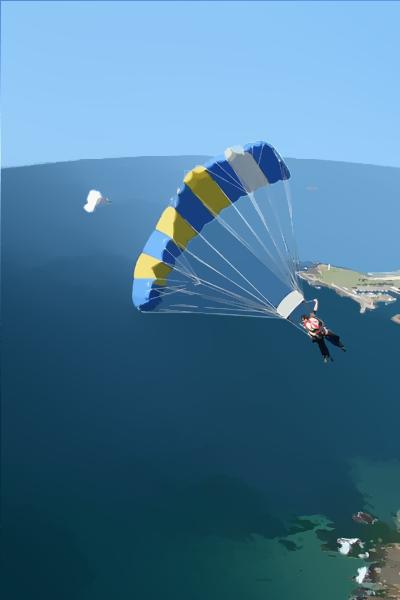}
&\includegraphics[width=2.9cm,trim={0 0 0 0},clip]{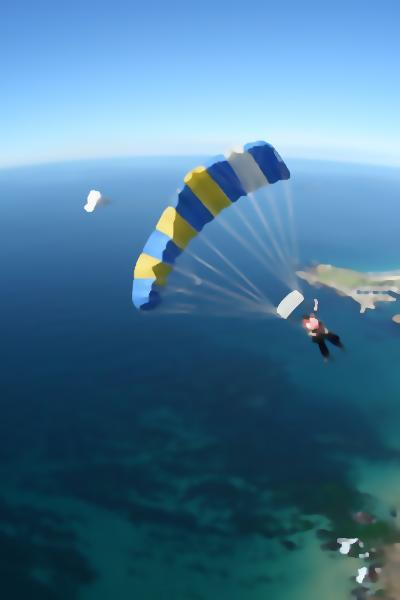}

\vspace{-1pt}\\

Input & Ours & SGF & SDF & $L_1$ & BTLF

\vspace{4pt}\\
\includegraphics[width=2.9cm,trim={0 0 0 0},clip]{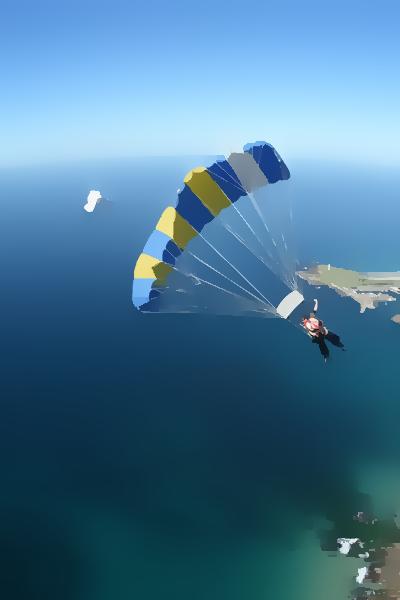}
&\includegraphics[width=2.9cm,trim={0 0 0 0},clip]{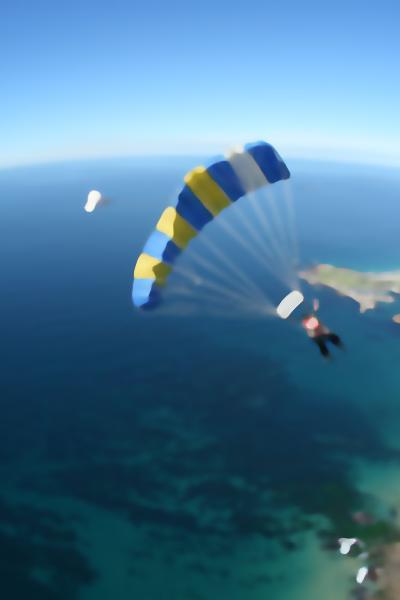}
&\includegraphics[width=2.9cm,trim={0 0 0 0},clip]{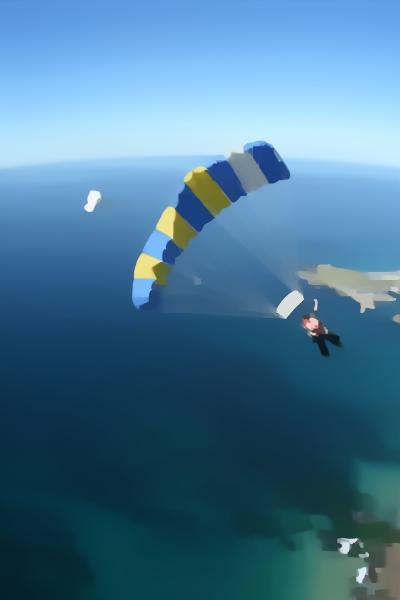}
&\includegraphics[width=2.9cm,trim={0 0 0 0},clip]{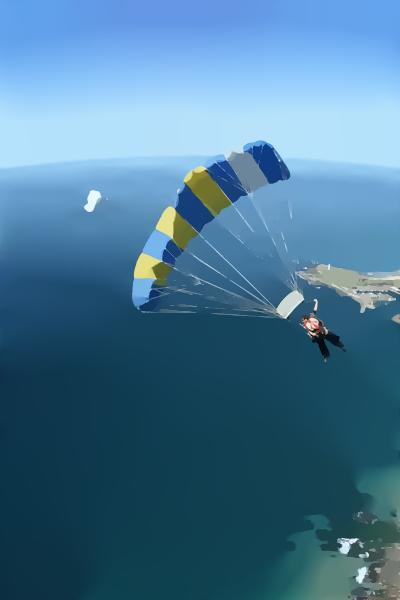}
&\includegraphics[width=2.9cm,trim={0 0 0 0},clip]{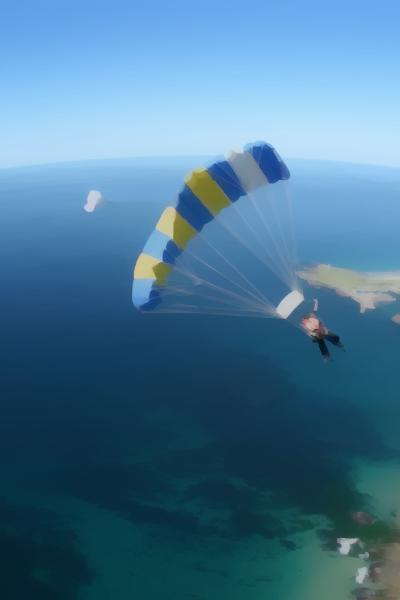}
&\includegraphics[width=2.9cm,trim={0 0 0 0},clip]{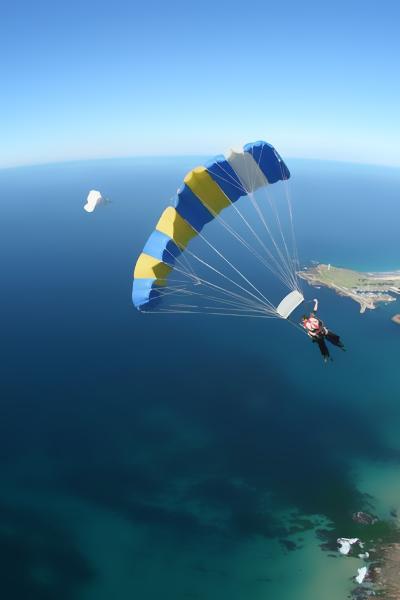}
\vspace{-1pt}\\

FGS & RGF & RTV & $L_0$ & WLS & BLF

%\includegraphics[width=2.9cm,trim={0 0 0 0},clip]{sig_input_47.jpg}
%&\includegraphics[width=2.9cm,trim={0 0 0 0},clip]{sig_input_47-predict_standard_edgePreserving_L2_weight5_window3_threshold_20_10_window10_L08_w1_30.jpg}
%&\includegraphics[width=2.9cm,trim={0 0 0 0},clip]{sig_input_47-SGF.jpg}
%&\includegraphics[width=2.9cm,trim={0 0 0 0},clip]{sig_input_47-SDF.jpg}
%&\includegraphics[width=2.9cm,trim={0 0 0 0},clip]{sig_input_47-L1smooth.jpg}
%&\includegraphics[width=2.9cm,trim={0 0 0 0},clip]{sig_input_47-BTLF.jpg}
%
%\vspace{-1pt}\\
%
%Input & Ours & SGF & SDF & $L_1$ & BTLF
%
%\vspace{4pt}\\
%\includegraphics[width=2.9cm,trim={0 0 0 0},clip]{sig_input_47-FGS.jpg}
%&\includegraphics[width=2.9cm,trim={0 0 0 0},clip]{sig_input_47-RGF.jpg}
%&\includegraphics[width=2.9cm,trim={0 0 0 0},clip]{sig_input_47-RTV.jpg}
%&\includegraphics[width=2.9cm,trim={0 0 0 0},clip]{sig_input_47-L0smooth.jpg}
%&\includegraphics[width=2.9cm,trim={0 0 0 0},clip]{sig_input_47-WLS.jpg}
%&\includegraphics[width=2.9cm,trim={0 0 0 0},clip]{sig_input_47-BLF.jpg}
%\vspace{-1pt}\\
%
%FGS & RGF & RTV & $L_0$ & WLS & BLF
\end{tabular}

\end{center}
\vspace{-3mm}
\caption{Visual comparison between our method and previous image smoothing methods, abbreviated as SGF~\cite{zhang2015segment}, SDF~\cite{ham2015robust}, $L_1$~\cite{L1smooth}, BTLF~\cite{cho2014bilateral}, FGS~\cite{FGS}, RGF~\cite{RGF}, RTV~\cite{RTV}, $L_0$~\cite{L0smooth}, WLS~\cite{WLS} and BLF~\cite{BLF98}. Our smooth image is generated by depressing the low-amplitude details and preserve the high-contrast structures. It can be seen that in addition to achieving pleasing flattening effects, the slender ropes of the parachute are also maintained much better in our result than the others. \textbf{Zoom in} to see the details. Photo courtesy of Skydive Australia.}
%Photo courtesy of Flickr user Sheba\_Also.
\label{figure:comp_others}
\vspace{-0mm}
\end{figure*}

\section{Related Work}

As a fundamental tool for many computer vision and graphics applications, image smoothing has been extensively studied in the past decades. The filtering based approaches, such as anisotropic diffusion \cite{anisotropic}, bilateral filter \cite{BLF98} and many others \cite{weiss2006fast,paris2006fast,fattal2009edge,kass2010smoothed,DT,chen2007real} have been the dominating image smoothing solutions for a long time. The core idea for such methods lies in filtering each pixel with its local spatial neighbourhood, and these methods are usually very efficient.

Recently, algorithms using mathematical optimization for image smoothing tasks gain more popularity due to their robustness, flexibility and more importantly the superiority of the smoothing results.
For example, \cite{WLS} proposed an edge-preserving operator in a weighted least square (WLS) optimization framework, which prevents the local image regions from being over-sharpened with an $L_2$ norm. Similar schemes have been achieved more efficiently by \cite{FGS,liu2017semi}. These works are devoted to extracting and manipulating the image details using image smoothing for various applications such as detail enhancement, HDR tone mapping, \emph{etc}.

On the other hand, \cite{L0smooth} proposed a sparse gradient counting scheme in an optimization framework by minimizing the $L_0$ norm. The method of \cite{L1smooth} aimed at producing almost ideally flattening images where sharp edges are also well preserved with $L_1$ norm. These methods are particularly well-suited for preserving or enhancing the sharp edges, and remove the low-amplitude details. They can be useful for some stylization effects or intrinsic image decompositions.

In the aforementioned applications, the image smoothing algorithms typically exploit only gradient magnitude as the main cue to discriminate primary image structures from details. \cite{RTV} presented the specifically designed relative total variation measures to extract meaningful structure, and \cite{ham2015robust} fuses appropriate structures of static and dynamic guidance images into a non-convex regularizer. Their goal is to remove fine-scale repetitive textures where local gradient can still be significant, which can not be easily achieved by the aforementioned smoothing approaches.

This regularization idea can also be interpreted as an image prior formulated in a deep network \cite{ulyanov2017deep} or image denoising engine \cite{romano2017little}. Discussions about the $L_p$-norm regularization can also be found in \cite{prasath2015multiscale,chung2009image,bach2012structured}. Interestingly, \cite{mrazek2006robust} observe that even the image filters can all be derived from minimization of a single energy functional with data and smoothness term.

Most of the optimization based approaches are time-consuming, as they typically require solving large-scale linear systems (or others). Therefore, recently some methods such as \cite{xu2015,liu2016,fan2017generic,chen2017fast,gharbi2017deep} were proposed to speed up existing smoothing operators. These methods train a deep neural network using the ground-truth smoothed images generated by existing smoothing algorithms. In contrast, our neural network is trained by optimizing an objective function through deep neural network in an unsupervised fashion. Note these previous deep models and ours are fundamentally different in many fields: target goal, training data, essential algorithm logic, \emph{etc.} Since they aimed at approximating traditional image smoothing algorithms, while ours creates some novel and unique smoothing effects, it makes our results not directly comparable to theirs by the quality of smooth images.

Deep learning has been applied to many image manipulation tasks \cite{chen2017stylebank,fan2018revisiting,he2018deep,chen2017coherent,chen2018stereoscopic}. But most previous work treat deep learning as a regression or classification tool. In this paper, we apply deep neural network as an optimization solution in a label-free setup.
% and analyze its performance in optimizing objective functions with both numerical and visual comparisons.

\section{Approach} \label{sec:approach}
In this section, we introduce our proposed formulation, including an edge-preserving criterion in Section~\ref{sec:wholecriterion} and a spatially adaptive $L_p$ flattening criterion in Section~\ref{sec:spatiallyvariant} which account for structure preservation and detail elimination respectively. Later on we describe how deep learning is leveraged for optimizing the proposed objective in Section~\ref{sec:networkstructure}.

\subsection{Objective Function Definition}\label{sec:wholecriterion}
Image smoothing aims at diminishing unimportant image details while maintaining primary image structures. To achieve this using energy minimization, our overall energy function for image smoothing is formulated as
\begin{equation}\label{eq:energy}
\mathcal{E} = \mathcal{E}_d + \lambda_f \cdot \mathcal{E}_f + \lambda_e \cdot \mathcal{E}_e,
\end{equation}
where $\mathcal{E}_d$ is the data term, $\mathcal{E}_f$ is the regularization term and $\mathcal{E}_e$ is the edge-preserving term. $\lambda_f$ and $\lambda_e$ are constant balancing weights.

The data term minimizes the difference between the input image and the smoothed image to ensure structure similarity. Denoting the input image by $I$ and the output image by $T$, both in RGB color space, a simple data term can be defined as
\begin{equation}\label{eq:Ed}
\mathcal{E}_d = \frac{1}{N} \sum\limits_{i=1}^N ||T_i-I_i||_2^2,
\end{equation}
where $i$ denotes pixel index and $N$ is the total pixel number.

Some important edges may be missed or weakened during the smoothing process since the goal of color flattening naturally conflicts with edge preserving to some extent. To address this issue, we propose an explicit edge-preserving criterion which preserves important edge pixels.

Before presenting this criterion, we first introduce the concept of \emph{guidance image}, which is formulated as the edge response of an image in appearance. A simple form of edge response is the local gradient magnitude:
\begin{equation}\label{eq:Edge_forward}
E_i(I) = \sum_{j \in \mathcal{N}(i)} | \sum_c(I_{i,c}-I_{j,c}) |
\end{equation}
where $\mathcal{N}(i)$ denotes the neighborhoods of point $i$ and $c$ denotes the color channel of the input image $I$. A similar guidance edge map of the output smooth image $T$ can also be calculated as $E(T)$.

Our edge-preserving criterion is defined by minimizing the quadratic difference of their edge responses between the guidance edge images $E(I)$ and $E(T)$. Let $B$ be an binary map where $B_i=1$ indicates an important edge point and $0$ otherwise, our edge-preserving term is defined as
\begin{equation}\label{Pi}
\mathcal{E}_e = \frac{1}{N_e} \sum\limits_{i=1}^{N}B_i\cdot ||E_i(T)-E_i(I)||_2^2
\end{equation}
where $N_e=\sum_{i=1}^{N}B_i$ is the total number of important edge points.

The definition of ``important edges" is more subjective and varies across different applications. The ideal way to obtain binary maps $B$ would be manual labeling with user preference. However, pixel-level manual labeling is rather labor-intensive. In this paper, we leverage a heuristic yet effective method to detect edges. Since this process is not our main contribution, we defer the detailed description of this edge detector to the supplemental material. A few examples of the detected major image structure are shown in Section~\ref{abstraction}. Also note that any previous advanced and sophisticated edge detection algorithm can all be certainly incorporated based on user preference. 

\begin{figure}[t]
\begin{center}
\begin{tabular}{ccc}

%\includegraphics[width=2.8cm,trim={0 0 0 0},clip]{demo7-22-input.jpg}
%&\includegraphics[width=2.8cm,trim={0 0 0 0},clip]{demo7-22-input-predict_standard_selfcomp_edgePreserving_woFlatten_l2only_window10_L08_w1_epoch30.jpg}
%&\includegraphics[width=2.8cm,trim={0 0 0 0},clip]{demo7-22-input-predict_standard_edgePreserving_L2_weight5_window3_threshold_20_10_window10_L08_w1_epoch30.jpg}

\includegraphics[width=2.8cm,trim={0 0 0 0},clip]{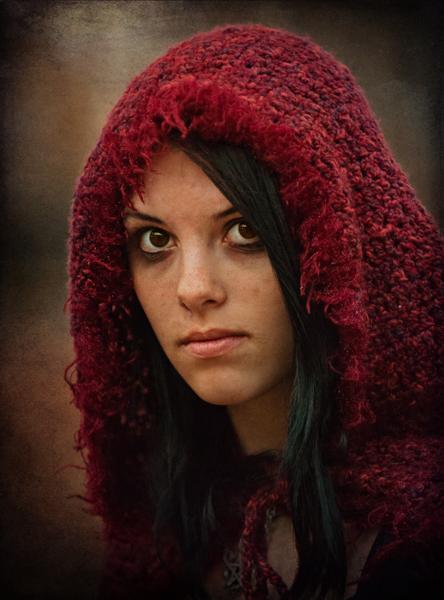}
&\includegraphics[width=2.8cm,trim={0 0 0 0},clip]{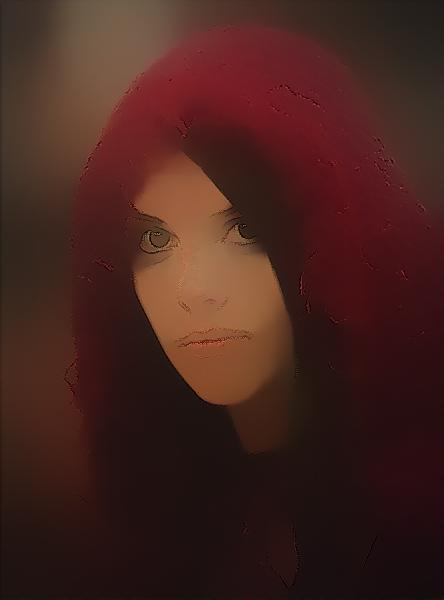}
&\includegraphics[width=2.8cm,trim={0 0 0 0},clip]{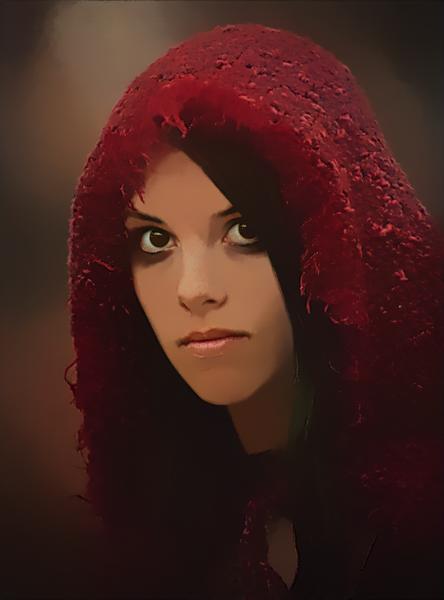}
\vspace{-2pt}\\
\small{Input} & \small{w/o $L_{0.8}$ norm} & \small{Ours} \vspace{3pt}\\

%\includegraphics[width=2.8cm,trim={0 0 0 0},clip]{demo-new2-17_vis.jpg}
%&\includegraphics[width=2.8cm,trim={0 0 0 0},clip]{demo-new2-17-predict_standard_selfcomp_edgePreserving_woFlatten_l08only_window10_L08_w1_epoch30_vis.jpg}
%&\includegraphics[width=2.8cm,trim={0 0 0 0},clip]{demo-new2-17-predict_standard_edgePreserving_L2_weight5_window3_threshold_20_10_window10_L08_w1_epoch30_vis.jpg}

\includegraphics[width=2.8cm,trim={0 2cm 0 0},clip]{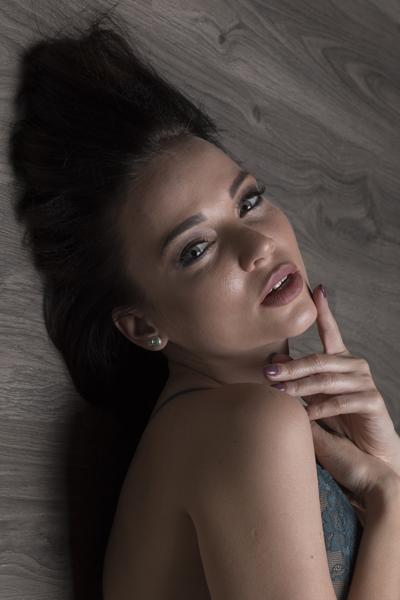}
&\includegraphics[width=2.8cm,trim={0 2cm 0 0},clip]{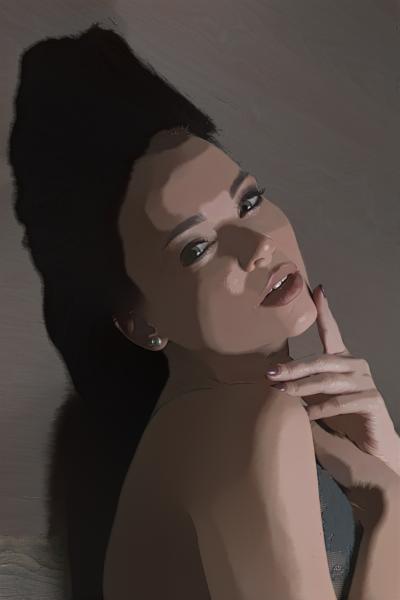}
&\includegraphics[width=2.8cm,trim={0 2cm 0 0},clip]{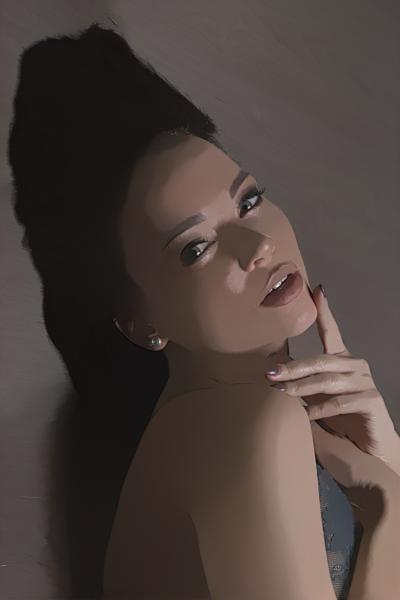}
\vspace{-2pt}\\
\small{Input} & \small{w/o $L_{2}$ norm} & \small{Ours}
\vspace{3pt}\\

%\includegraphics[width=2.8cm,trim={0 0 0 0},clip]{demo9-400-1-input.jpg}
%&\includegraphics[width=2.8cm,trim={0 0 0 0},clip]{demo9-400-1-input-predict_standard_selfcomp_edgePreserving_L2_weight5_window3_threshold_20_10_window10_L08_w1_woEdge_epoch30.jpg}
%&\includegraphics[width=2.8cm,trim={0 0 0 0},clip]{demo9-400-1-input-predict_standard_edgePreserving_L2_weight5_window3_threshold_20_10_window10_L08_w1_epoch30.jpg}

\includegraphics[width=2.8cm,trim={0 0 0 0},clip]{sig_input_47.jpg}
&\includegraphics[width=2.8cm,trim={0 0 0 0},clip]{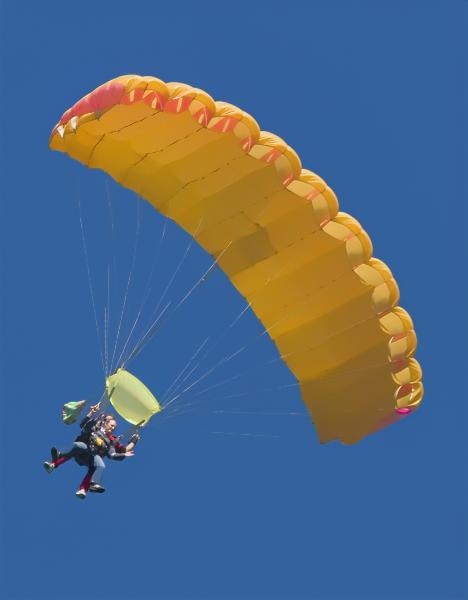}
&\includegraphics[width=2.8cm,trim={0 0 0 0},clip]{sig_input_47-predict_standard_edgePreserving_L2_weight5_window3_threshold_20_10_window10_L08_w1_30.jpg}

\vspace{-2pt}\\
\small{Input} & \small{w/o EP criterion} & \small{Ours}
\\

\end{tabular}
\end{center}
\vspace{-3mm}
\caption{The effectiveness of our proposed criterions. Note each individual part is essential to generate our visually-pleasing smooth results. \textbf{Zoom in} to see the details. Photo courtesy of Flickr user Michael Miller, Andre Wislez and  Sheba\_Also.}
\label{figure:comp_lp}
\vspace{0mm}
\end{figure}

Given sufficient training images with classified edge points, the deep network will implicitly learn the edge importance through minimizing the edge-preserving term and reflect such information in the smooth images.
Figure~\ref{figure:comp_lp} demonstrates an extremely difficult case of a parachute. As can be seen, without the edge-preserving criterion, some thin yet semantically important structures like the white rope are smoothed out in the output images. In contrast, the result optimized with our full criterions maintains these structures very well.

\setlength{\tabcolsep}{1pt}
\begin{figure*}[htp]
	\centering
	\includegraphics[scale=0.267]{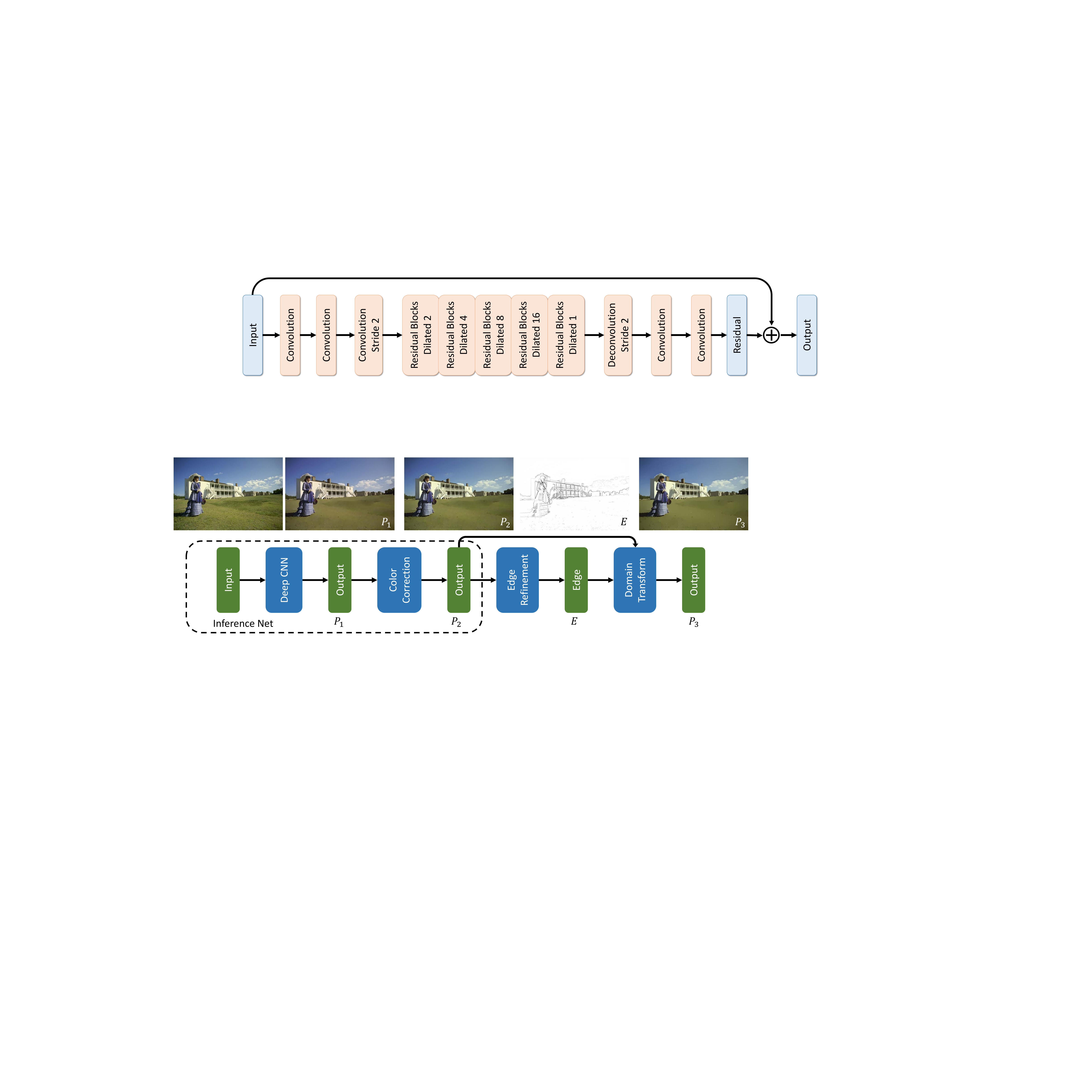}
	\caption{The network structure used throughout this paper. Our network contains 26 convolution layers where the middle 20 layers are organized into residual blocks of dilated convolutions with exponentially increasing dilation factor (except for the last residual block) to enlarge the receptive field. The skip connection from the input to the output is employed such that the network learns to predict a residual image.}
	\label{figure:network}
	%\vspace{-4mm}
\end{figure*}

\subsection{Dynamic Spatially-Variant $L_p$ Flattening Criterion} \label{sec:spatiallyvariant}

We now present our new smoothness/flattening term with spatially-variant $L_p$ norms on the image in order to gain better quality and more flexibility.

To remove unwanted image details, the smoothing or flattening term advocates smoothness for the solution by penalizing the color differences between adjacent pixels:
\begin{equation}
\begin{aligned}\label{eq:Ef2}
\mathcal{E}_f = \frac{1}{N} \sum\limits_{i=1}^N \sum_{j \in \mathcal{N}_h(i)} w_{i,j} \cdot |T_i-T_{j}|^{p_i}, \\
\end{aligned}
\end{equation}
where $\mathcal{N}_h(i)$ denotes the adjacent pixels of $i$ in its $h \times h$ window, $w_{i,j}$ denotes the weight for the pixel pairs and $|\cdot|^p$ is an $L_p$ norm\footnote{With slight abuse of terminology, we use $L_p$ norm to refer to the $L_p$ norm raised to the $p$-th power, \emph{i.e.}, it will indicate $(|\cdot|^p\!+\!\cdots\!+\!|\cdot|^p)$ as opposed to $(|\cdot|^p\!+\!\cdots\!+\!|\cdot|^p)^\frac{1}{p}$.}.

The weight $w_{i,j}$ can be calculated from either color affinity or spatial affinity (or their combination), which are defined respectively as
\begin{equation}\label{eq:wr}
w_{i,j}^r = \exp(-\frac{\sum_c(I_{i,c}-I_{j,c})^2}{2\sigma_r^2}),
\end{equation}
\begin{equation}\label{eq:ws}
w_{i,j}^s = \exp(-\frac{(x_i - x_j)^2 + (y_i - y_j)^2}{2\sigma_s^2}),
\end{equation}
where $\sigma_r$ and $\sigma_s$ are the standard deviations for the Gaussian kernels computed in either color space or spatial space, $c$ denotes image channel (in this paper we use the YUV color space to compute weights $w_{i,j}^r$ in Equation~\ref{eq:wr}), and $x,y$ denote pixel coordinates.

Determining the image regions for different $L_p$ regularizers is not trivial. To help locate these regions in our algorithm, we leverage the guidance images to define the value of $p_i$ and its corresponding weight for each image pixel as
\begin{equation}
\begin{aligned}\label{eq:p}
p_i,\ w_{i,j} =
\begin{cases}
p^{large},\ w_{i,j}^s &\text{if } E_i(I) < c_1 \text{ and } E_i(T) - E_i(I) > c_2,\\
p^{small},\ w_{i,j}^r &\mbox{otherwise}.
\end{cases}
\end{aligned}
\end{equation}
where $p^{large}$ and $p^{small}$ are two representative values for $p$, and $c_1$ and $c_2$ denote two positive thresholds. We set $p^{large}=2$ and $p^{small}=0.8$ throughout this paper. It can be seen that the $p$ value distribution is not determined \emph{a priori} with the input image, but is conditioned on the output image. We explain such a strategy in the following two points.

\paragraph{Suppressing artifacts caused by single regularizer} The intuition behind Equation~\ref{eq:p} is that when we minimize the energy function, $L_{0.8}$ norm is applied until some over-sharpened spurious structures appear in the output image due to the piecewise constant effect caused by $L_{0.8}$ regularizer, at which time $L_2$ norm will be applied to suppress the artifact. These spurious structures are identified as the ones whose edge response of pixel $i$ on the original image $I$ is low (as characterized by $E_i(I) < c_1$) but is significantly heightened on the output image $T$ (per $E_i(T) - E_i(i) > c_2$). In Figure~\ref{figure:comp_lp}, we demonstrate a few smooth results optimized with our objective function. As can be seen, without $L_2$ norm, it achieves strong smoothing effects but also yields staircasing artifacts on the lady's cheek and shoulder. On the other hand, without $L_{0.8}$ norm, the optimized image is very blurry and many important structures are not well preserved due to the $L_2$ regularizer. On the contrary, the results optimized with our proposed full criterions are much more visually pleasing.

\paragraph{Enabling different applications via specialized guidance images} Our guidance image and spatially-variant $L_p$ flattening norm also enable us to achieve flexible smoothing effects for different applications. For example, if the goal is to remove a certain type of image structures like small-scale textures, we can simply eliminate all the edge points belonging to these textures in the guidance image $E(I)$ by setting their values to zero. This way, the edge responses $E(T)$ on these regions of the output image will always be larger, and $L_2$ norm will be applied to remove these textures. Later we show two such applications -- texture removal (Section~\ref{sec:textureremoval}) and content-aware image manipulation (Section~\ref{sec:contentaware}). All other results shown in this paper are obtained by raw guidance images computed via Equation~\ref{eq:Edge_forward}.

Note that we adopt spatial affinity to calculate the weights $w_s$ for regions with an $L_2$ norm, as it is more effective for edge suppression. Color affinity is utilized for $L_{0.8}$ norm regions for better flattening effect. Since $L_2$ and $L_{0.8}$ norm regularize the images differently, we amplifies the weight of $L_2$ norm with a scale scalar $\alpha$ for balance. We empirically determine these two $p$ values, which represent the regularization for strong flattening and blurring effects in a general sense. They are replaceable with other alternatives.

It can be seen that our spatially variant $L_p$ norm is not fixed, but \emph{dynamically changing} in the iterative optimization (training) procedure based on the output image. Although we do not provide a theoretical proof of convergence, we have found empirically that such a procedure converges and the $p$ value distribution stabilizes in the end. Note we observe that \cite{zhu2014spectral} also employs data-guided sparsity in their work, but differently their regularization is static while ours is dynamically changed from the output image.

%In summary, with the dynamic spatially-variant $L_p$ flattening term, one can not only suppress the staircasing artifacts (spurious edges) created by an individual sparsity-inducing regularizer and obtain pleasing flattening effect, but also preserve specific image structures in accordance to the different applications.

\subsection{Deep Learning based Optimization}\label{sec:networkstructure}
As the whole objective function is derivative to the optimized smooth image, we implement it as the loss layer in a deep learning framework. The loss function is optimized with gradient descent method through a deep neural network.
%Since there is no ground truth label for supervision, 
The whole training process is in an unsupervised learning fashion with a large number of unlabeled natural images. The deep network implicitly learns the optimization procedure and once the network is trained, it only requires one forward pass through the deep neural network to predict the smooth image without further optimization steps.

Now we introduce architecture of our deep neural network which is used for minimizing the defined energy function. 
Inspired by the previous work \cite{dilate} which enlarges the receptive field with dilated convolutions for semantic segmentation, and \cite{resolution} which uses very deep convolutional neural network equipped with residual learning for super-resolution, we design a fully convolutional network (FCN) equipped with dilated convolution and skip connections for our task.

\paragraph{Basic structure description} Figure~\ref{figure:network} is a schematic description of our FCN. The network contains 26 convolution layers, all of which use 3$\times$3 convolution kernels and outputs 64 feature maps (except for the last one which produces a 3-channel image). All the convolution operations are followed by batch normalization~\cite{ioffe2015batch} and ReLU activation except for the last layer. The third conv layer downsamples the feature maps by half via using a stride of 2, and the third from last layer is a deconvolution (aka fractionally-strided convolution) layer recovering the original image size. The middle 20 conv layers are organized as 10 residual blocks \cite{he2016deep}.
A full description of the detailed network structure is presented in the supplemental material.

\paragraph{Large receptive field in dilated convolutions} As image smoothing requires contextual information across wide regions, we increase the receptive field of our FCN by using dilated convolution with \emph{exponentially increasing dilation factors} except for the last residual block, similar to \cite{dilate}. Specifically, any two consecutive residual blocks share one dilation factor, which is doubled in the next two residual blocks. This is an effective and efficient strategy to increase receptive field without sacrificing image resolution: with exponentially increasing dilation factors, all the points in an $n \times n$ grid can be reached from any location in logarithmic steps $log(n)$. Similar strategies have been used in parallel GPU implementation of some traditional algorithms, such as Voronoi diagram~\cite{jumpflooding} and PatchMatch~\cite{fan2015jumpcut}.

\paragraph{Residual image learning}
In the image smoothing task, the input and output images are highly correlated. In order to ease the learning, instead of directly predicting a smoothed image we predict a \emph{residual image} and generate the final result via point-wise summation of the residual image and the raw input image. Such a residual image learning design avoids the color attenuation issue observed in previous works (\emph{e.g.}, \cite{resolution}). 

\section{Implementation Details}\label{sec:implementation}

Our FCN network and energy function are implemented in the Torch framework and optimized with mini-batch gradient descent. The batch size is set as 1. The network weights are randomly initialized using the method of \cite{delve}. The Adam \cite{Adam} algorithm is used for training with the learning rate set as 0.01. We train the network for 30 epoches, which takes about 8 hours on an NVIDIA Geforce 1080 GPU.

\paragraph{Training and testing data}Since our network does not require ground-truth smooth image for training, any image can be used to train it.
For better generalization to natural images, we use the PASCAL VOC dataset~\cite{everingham2010pascal} which contains about 17,000 images to train the network. These images were collected in the Flicker photo-sharing website and exhibit a wide range of scenes under a large variety of viewing conditions. We crop the images to the size of 224$\times$224 to accelerate the training process without jeopardizing the smoothing quality. Once the network is trained, we run it on images outside of PASCAL VOC and evaluate the results.

\paragraph{Parameter specifics:} The parameters in our proposed objective function are specified by default as follows: 0.1 ($\sigma_r$), 7 ($\sigma_s$), 1 ($\lambda_f$), 0.1 ($\lambda_e$), 5 ($\alpha$), 20 ($c_1$), 10 ($c_1$), 21 ($h$). To achieve the optimal performance on each individual application, a small subset of parameters may be tweaked, which is discussed in Section \ref{sec:applications}. However, note that these parameters are only tuned based on the application type, not on any particular images. All the images in the same application shown in both the paper and supplemental material are generated by the same set of parameter values.
%, which demonstrates the robustness of our algorithm.

\section{method analysis and discussion}

\setlength{\tabcolsep}{1pt}
\begin{figure}[t]
\begin{center}
\begin{tabular}{cccc}
%\includegraphics[width=2.1cm,trim={0 0 0 0},clip]{demo-new2-17.jpg}
%&\includegraphics[width=2.1cm,trim={0 0 0 0},clip]{demo-new2-17-adam_spatailGau_100.jpg}
%&\includegraphics[width=2.1cm,trim={0 0 0 0},clip]{demo-new2-17-IRLS-10-small-new.jpg}
%&\includegraphics[width=2.1cm,trim={0 0 0 0},clip]{demo-new2-17-predict_standard_selfcomp_edgePreserving_L2_weight3_window4_threshold_20_10_woEdge_epoch25.jpg}
%\\

\includegraphics[width=2.1cm,trim={0 0 0 0},clip]{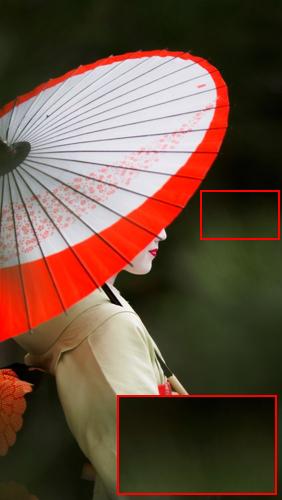}
&\includegraphics[width=2.1cm,trim={0 0 0 0},clip]{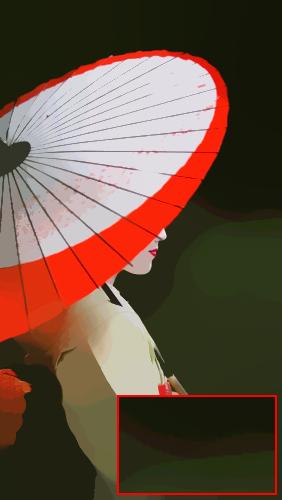}
&\includegraphics[width=2.1cm,trim={0 0 0 0},clip]{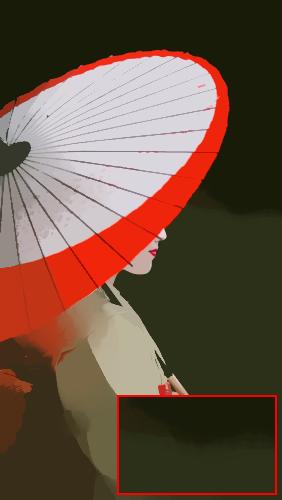}
&\includegraphics[width=2.1cm,trim={0 0 0 0},clip]{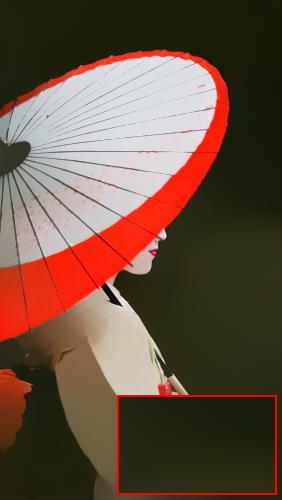}
\\

\includegraphics[width=2.1cm,trim={0 0 0 0},clip]{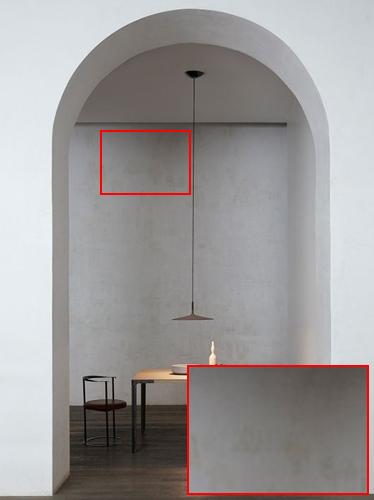}
&\includegraphics[width=2.1cm,trim={0 0 0 0},clip]{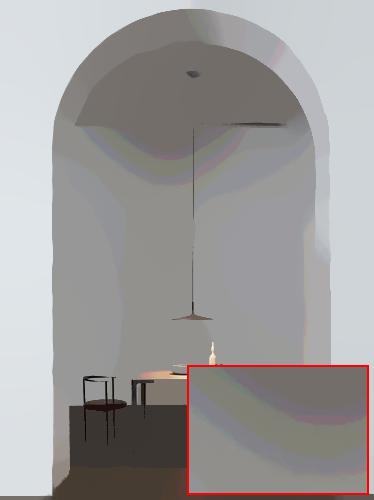}
&\includegraphics[width=2.1cm,trim={0 0 0 0},clip]{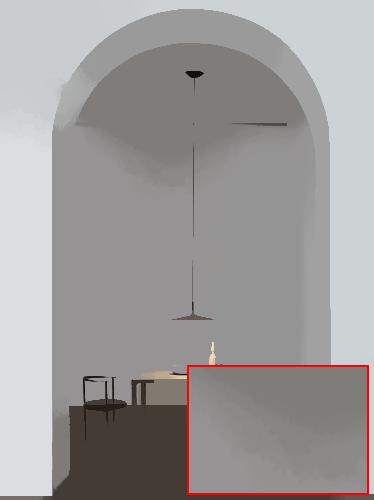}
&\includegraphics[width=2.1cm,trim={0 0 0 0},clip]{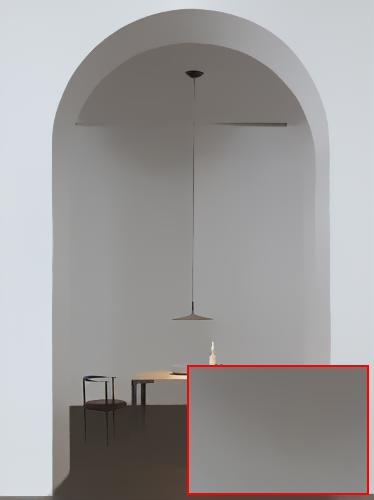}
\\

%\includegraphics[width=2.1cm,trim={0 0 0 0},clip]{demo-new5-4.jpg}
%&\includegraphics[width=2.1cm,trim={0 0 0 0},clip]{demo-new5-4-adam_spatailGau_100.jpg}
%&\includegraphics[width=2.1cm,trim={0 0 0 0},clip]{demo-new5-4-IRLS-10-small-new.jpg}
%&\includegraphics[width=2.1cm,trim={0 0 0 0},clip]{demo-new5-4-predict_standard_selfcomp_edgePreserving_L2_weight5_window3_threshold_20_10_window10_L08_w1_woEdge_epoch28.jpg}
%\\
\small{Input} & \small{Adam} & \small{IRLS} & \small{Ours}
\\

\end{tabular}
\end{center}
\vspace{-3mm}
\caption{Comparison between our learned deep neural network and traditional numerical solvers. Compared with the gradient descent optimizer Adam \cite{Adam} and Iterative Reweighted Least Square (IRLS) \cite{holland1977robust}, our results are visually more pleasing which do not have the spurious staircasing structures. Photo courtesy of Tumblr user gaaplite and LucidiPevere Studio.}
\label{figure:comp_network}
\vspace{-3mm}
\end{figure}

In this section, we first compare the smooth images optimized with our deep learning solver and traditional numerical solvers, followed by the analysis of convergence of different optimizers and the potential reason why the deep learning solver achieves more visually pleasing results than the others for our problem.

\setlength{\tabcolsep}{2pt}
\begin{figure*}[htp]
\begin{center}
\begin{tabular}{ccc}

\small{$L_{2}$} &  \small{$L_{0.8}$} & \small{half $L_{2}$, half $L_{0.8}$}
\\

\includegraphics[width=5.9cm,trim={0cm 0cm 0cm 0cm},clip]{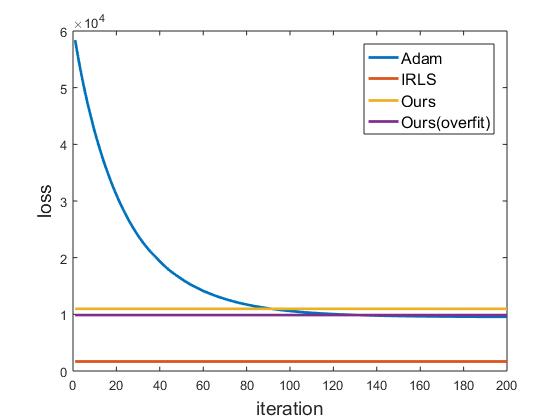}
&\includegraphics[width=5.9cm,trim={0cm 0cm 0cm 0cm},clip]{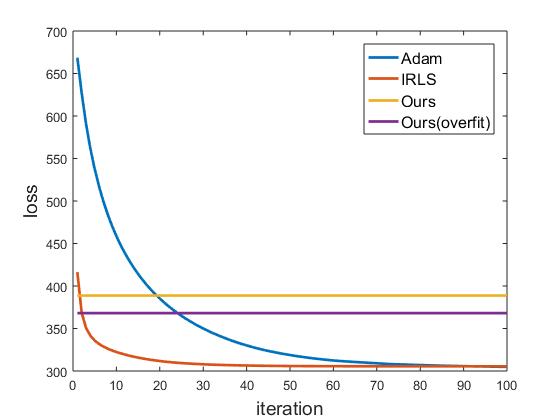}
&\includegraphics[width=5.9cm,trim={0cm 0cm 0cm 0cm},clip]{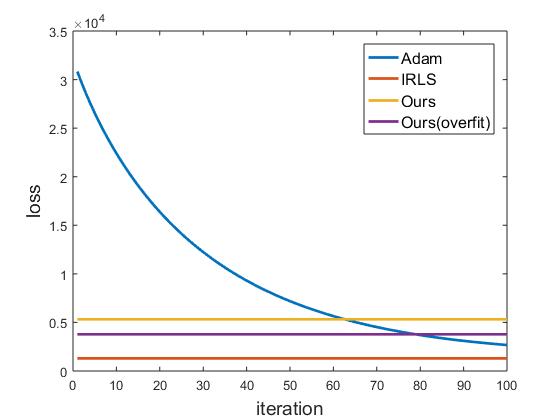}
\\

\end{tabular}
\end{center}
\vspace{-4mm}
\caption{The loss values from four optimization solvers (IRLS, Adam, our deep learning solver and the deep learning solver overfitted on a single image) for objective functions that contain either $L_2$ norm, $L_{0.8}$ norm or mixed of these two in the flattening criterion. hNote the loss is computed in the evaluation stage; our deep learning solver is non-iterative thus the loss is constant (as represented by a horizontal line).}
\label{figure:loss_curve}
\vspace{1mm}
\end{figure*}

\setlength{\tabcolsep}{1pt}
\begin{figure}[htp]
\begin{center}
\begin{tabular}{ccccc}

\raisebox{0.7\height}{\rotatebox{90}{{{$L_2$ norm}}}}
&\includegraphics[width=1.98cm,trim={0cm 0cm 0cm 0},clip]{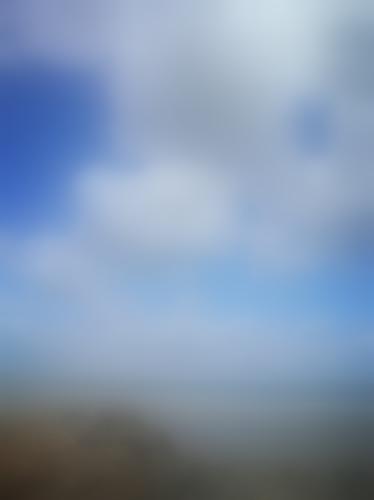}
&\includegraphics[width=1.98cm,trim={0cm 0cm 0cm 0},clip]{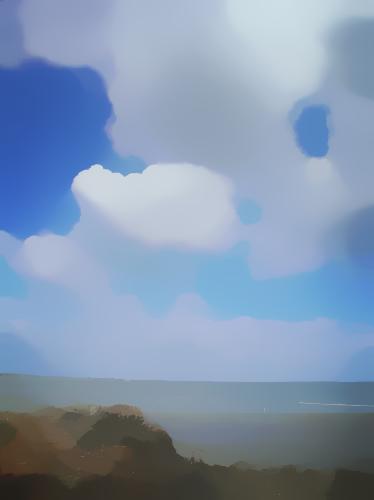}
&\includegraphics[width=1.98cm,trim={0cm 0cm 0cm 0},clip]{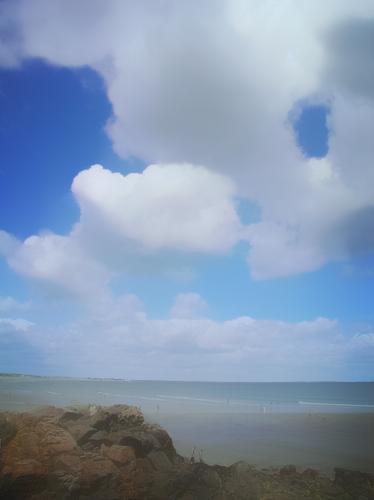}
&\includegraphics[width=1.98cm,trim={0cm 0cm 0cm 0},clip]{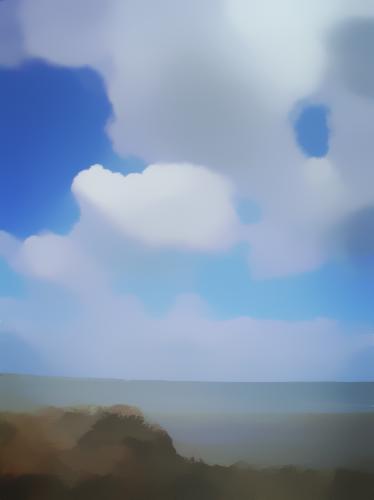}
\\

\raisebox{0.6\height}{\rotatebox{90}{{{$L_{0.8}$ norm}}}}
&\includegraphics[width=1.98cm,trim={0cm 0cm 0cm 0},clip]{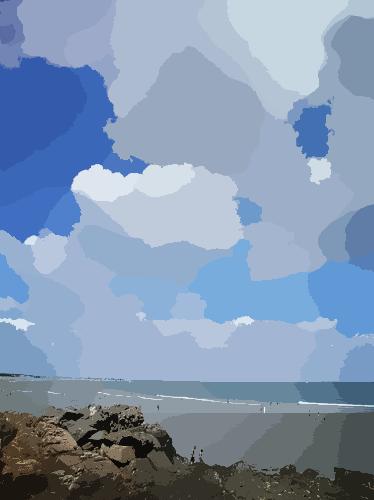}
&\includegraphics[width=1.98cm,trim={0cm 0cm 0cm 0},clip]{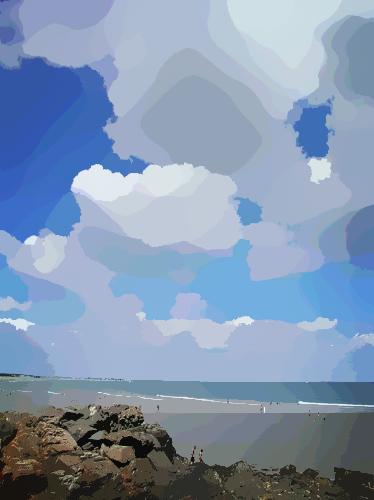}
&\includegraphics[width=1.98cm,trim={0cm 0cm 0cm 0},clip]{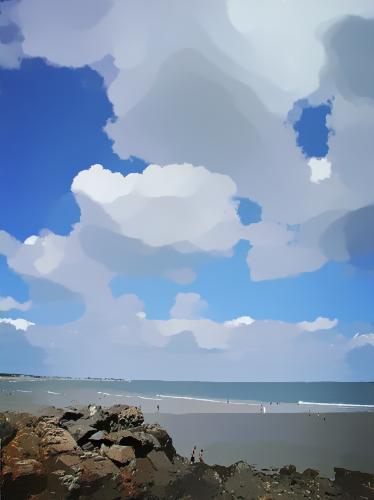}
&\includegraphics[width=1.98cm,trim={0cm 0cm 0cm 0},clip]{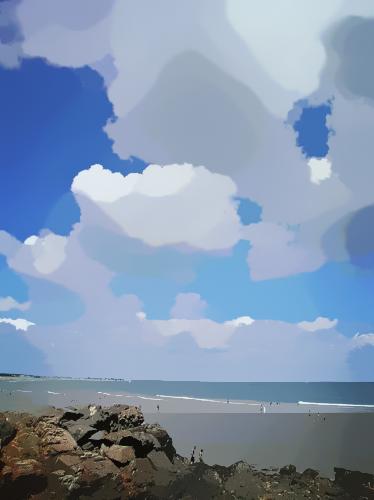}
\\

\raisebox{0.1\height}{\rotatebox{90}{{{half $L_2$, half $L_{0.8}$}}}}
&\includegraphics[width=1.98cm,trim={0cm 0cm 0cm 0},clip]{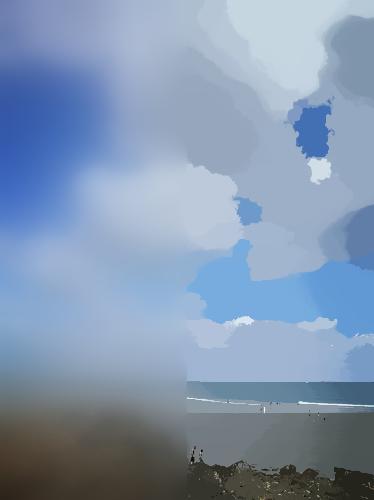}
&\includegraphics[width=1.98cm,trim={0cm 0cm 0cm 0},clip]{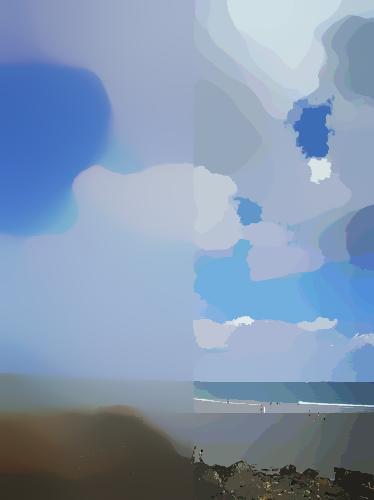}
&\includegraphics[width=1.98cm,trim={0cm 0cm 0cm 0},clip]{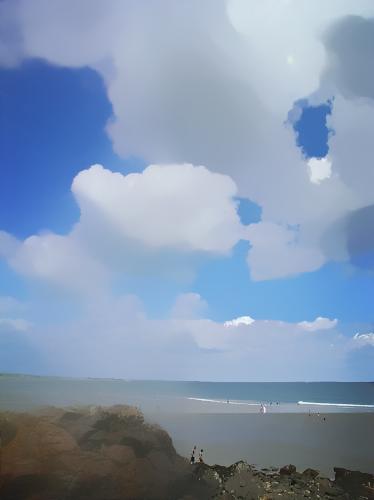}
&\includegraphics[width=1.98cm,trim={0cm 0cm 0cm 0},clip]{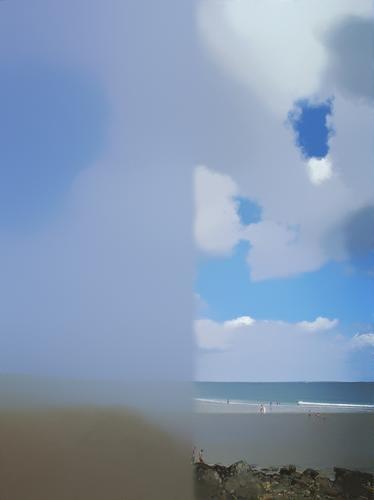}
\\

\raisebox{0.2\height}{\rotatebox{90}{{{adaptive norm}}}}
&\includegraphics[width=1.98cm,trim={0 0 0 0},clip]{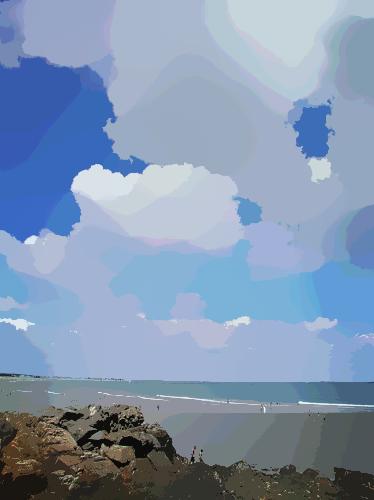}
&\includegraphics[width=1.98cm,trim={0 0 0 0},clip]{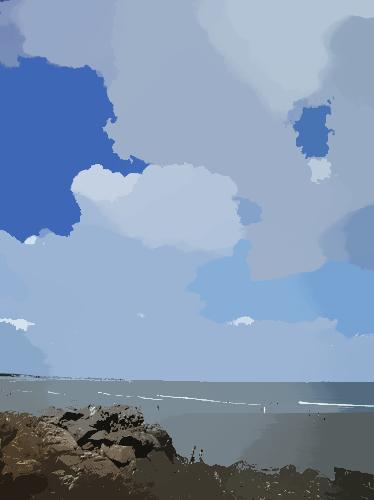}
&\includegraphics[width=1.98cm,trim={0 0 0 0},clip]{demo2-3-predict_standard_selfcomp_edgePreserving_L2_weight5_window3_threshold_20_10_window10_L08_w1_woEdge_epoch28.jpg}
&\includegraphics[width=1.98cm,trim={0 0 0 0},clip]{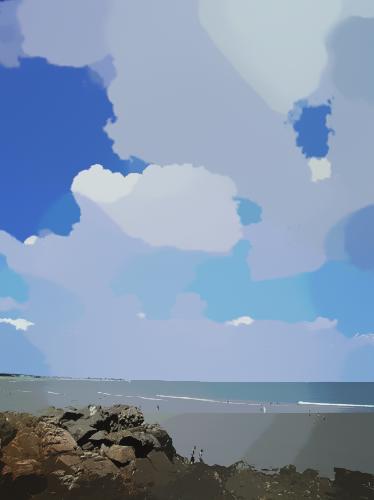}
\\

& IRLS & Adam & Ours & Ours (overfit)
\\

\end{tabular}
\end{center}
\vspace{-3mm}
\caption{Demonstration of the results optimized with $L_2$ norm, $L_{0.8}$ norm, half $L_2$ half $L_{0.8}$ and our adaptively-changed $L_p$ norms. \textbf{Zoom in} to see the detailed difference.  Photo courtesy of Flickr user Ryan L Collins.}
\label{figure:loss_images}
\vspace{0mm}
\end{figure}

\subsection{Visual results of different optimizers}
To verify the efficacy of our deep learning solver, as opposed to directly minimizing Equation \ref{eq:energy} using a traditional optimization algorithm, we compare it against two popular representative approaches, Adam and IRLS. Adam \cite{Adam} is a stochastic gradient descent-based optimization method that can be generically applied to nonconvex differentiable functions. Since our proposed objective is differentiable\footnote{Although $L_p$ norms are not differentiable on a set of measure zero, in such cases
subgradients can easily serve as a natural surrogate}, Adam is a very straightforward approach for minimizing it, at least locally. Likewise, iterative reweighted least square (IRLS) \cite{holland1977robust} represents a classical tool for minimizing energy functions with non-quadratic forms. Several image smoothing papers \cite{FGS,RTV} also employ IRLS allied with objectives regularized by $L_p$ norms ($0<p\leq1$).

To utilize IRLS for optimization, a tight quadratic upper bound of the energy has to be defined, which is trivial to accomplish for $L_p$ terms as has been done in the past. However, the proposed non-quadratic edge-preserving criterion cannot be bounded in this way, making IRLS problematic. Therefore for the results reported in this section, the energy function is formed from only the data term and the spatially adaptive $L_p$ flattening term for fair comparison across all three methods.  Smooth images optimized by different solvers are shown in Figure \ref{figure:comp_network}.  Note that Adam, as a gradient-based method requires 100 iterations to converge, while IRLS only requires about 10 iterations given that it applies second-order information in optimizing the quadratic upper bound.

In general, both the Adam and IRLS results are less satisfactory than our deep neural network solver.  For example, with Adam some spurious stair-casing edges are still generated as undesirable visual artifacts.  Likewise, for IRLS we observe over-sharpened side effects, although in places not quite as severe as with Adam. Also, the magnified local regions shown in the bottom of each figure display areas where the color intensity varies gradually, and both Adam and IRLS fail to smooth these areas well.
Also, the magnified local region in each figure shows that both Adam and IRLS cannot well smooth the local regions where color varies gradually.

%Since the optimized image is gradually changing in a small step each time, there form many independent stripes which break up the image into many local smooth regions, which can be clearly observed in the shown image.

\subsection{Performance analysis of different optimizers on fixed $L_p$ distribution}\label{sec:fixedlp}

Now we analyze the performance of these three optimizers by comparing their convergence curves. Note since our proposed objective function is adaptively changing based on the output images (per Equation~\ref{eq:Ef2} and ~\ref{eq:p}), it is not intuitive for comparing the convergence trend of these different optimizers. Thus we test three representative loss functions with fixed $L_p$ distribution map in $\mathcal{E}_f$, whose optimization difficulty is gradually increasing. They are the variants of our objective function, where we replace the adaptively changed regularizer with only $L_2$ norm, only $L_{0.8}$ norm, or fixed combination of these two.
%In the end, we discuss the relation between overfitting the deep learning solver over one single image and traditional solvers.
Following the previous ablation study, we disable the edge preserving term for IRLS. The loss values are averaged over 40 test images, and are shown in Figure~\ref{figure:loss_curve}. The corresponding visual results are shown in Figure~\ref{figure:loss_images}. Note these loss curves do not illustrate the training process of our method. Instead, they are constant values computed on the testing images.

%The leftmost figure in Figure~\ref{figure:loss_curve} shows the convergence of $L_2$ norm. Since the whole objective function becomes convex and quadratic with $L_2$ norm, IRLS is able to achieve the optimal results with one step. Adam is slightly better than our learned deep network, but the results of both methods are relatively far from optimal. Accordingly in Figure~\ref{figure:loss_images}, IRLS demonstrates the most blurry images regularized by $L_2$ norm.

With the fixed $L_2$ norm in $\mathcal{E}_f$, the whole objective function becomes convex and quadratic. Thus IRLS is able to achieve the optimal results with one step. Adam is slightly better than our learned deep network. However, the smoothing results of both methods are relatively far from optimal. Accordingly to Figure~\ref{figure:loss_images}, IRLS demonstrates the most blurry images regularized by $L_2$ norm.
When the objective function contains $L_{0.8}$ norm in $\mathcal{E}_f$, it becomes nonconvex. From the middle loss figure, we can see IRLS and Adam achieves similar energy value in the end, and the deep learning solver does not obtain a loss value as low as theirs. Figure~\ref{figure:loss_images} shows that the smoothing results of all the methods appear to be more piece-wise constant.
Finally, we demonstrate a case where the flattening criterion $\mathcal{E}_f$ contains $L_2$ norm in left half of the image and $L_{0.8}$ norm in right half. Different from our dynamically-changed $L_p$ norm, the $L_p$ distribution in this case is fixed for all different images and iterations.
%Its loss curve seems more like an average of the other two.
Figure~\ref{figure:loss_curve} shows that IRLS achieves the lowest energy value\footnote{To make the results more presentable, we slightly modified the objective function for IRLS and only show its final loss in this case.}, followed by Adam and deep learning solver.

\begin{figure}[t]
	\centering
	\includegraphics[scale=0.187]{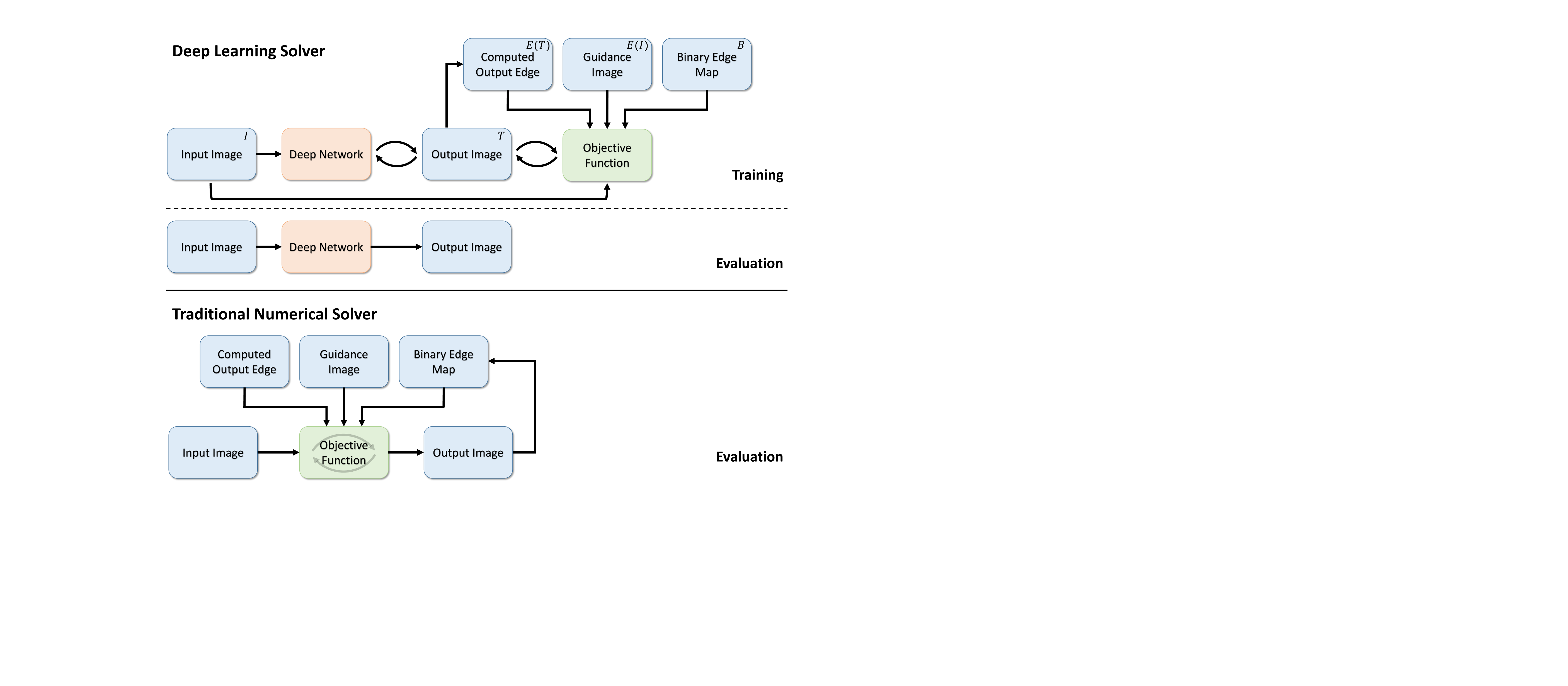}
	\caption{Workflow of our deep learning solver compared to traditional numerical solvers when applied to the proposed objective function. Our method employs a neural network to optimize an objective function with \emph{no ground truth labels} of smooth images. In the training stage, a few extra inputs are required to minimize the objective function; but once the network is train, the input image is only required to forward through the deep network once to predict smooth images in the evaluation stage. Regarding the traditional numerical solvers, given \emph{each new image}, all the different inputs are required to iteratively optimize the objective function.}
	\label{figure:workflow}
	\vspace{0mm}
\end{figure}

To understand why the traditional optimizers achieve lower energy values on the above three objective functions, we first illustrate the workflow of both traditional numerical solvers and our deep learning solver in Figure~\ref{figure:workflow}. As can be seen, given a particular image, traditional numerical solvers work by iteratively optimizing this single image. In contrast, the deep learning solver \emph{directly predicts their results from a one-forward-pass mapping, which is learned based on a large corpus of training data without any prior information on this specific image}. Thus traditional optimizers are more advantageous in achieving lower energy with fixed $L_p$ distribution, as verified by the above three loss curves.

To further analyze the above hypothesis, we overfit our network by training on only one single image, and compare both the final loss and the visual results. This way, the deep learning solver works similarly to traditional solvers. As can be seen, our deep learning solver with overfitting is able to achieve much lower energy, especially in the case of $L_2$ norm where our overfitting results are almost identical to Adam. Likewise, Figure~\ref{figure:loss_images} shows that the visual result obtained with the overfitting solver are visually much closer to IRLS and Adam. For example, there is a clear separation line between the two regularized regions for the half $L_2$ half $L_{0.8}$ case, which is not present in the results of the original deep learning solver.

%for the half $L_2$ half $L_{0.8}$ case, the result optimized with the overfitting solver appear similarly to IRLS and Adam which show a clear separation line between two regularized regions, however which effect isn't learned well by the original deep learning solver.

Note the loss functions defined in this subsection are used to analyze the performance of different solvers. They are not the actual loss function used for our image smoothing task.

%\vspace{-3mm}
\subsection{Performance analysis of different optimizers on our dynamic $L_p$ distribution}
In our proposed edge flattening criterion, the $L_p$ distribution is adaptively changing based on the output images. The whole objective function is highly complex and loss curve is not guaranteed to converge. Therefore, comparing the loss curves of the different solvers is less informative.
The last row of Figure~\ref{figure:loss_images} shows an additional set of results obtained under our adaptively changing $L_p$ norm. It can be observed the staircasing artifacts exist in the results of all IRLS, Adam, and the overfitting-based deep learning solver. In contrast, our deep learning solver generates very smooth results with no such artifacts.

%When it comes to the case with no fixed $L_p$ norm distribution like the proposed dynamically changed objective function, the \emph{loss curve is not guaranteed to converge}. Comparing the convergence of the three solvers on objective functions which contain different $L_p$ norm distributions is not fair enough and especially informative. Thus we wonder what should be the best results by visual perception. We show such an additional set of results in Figure~\ref{figure:loss_images}, as can be seen, the staircasing artifacts exist in all IRLS, Adam and the overfit deep learning solver. By contrast, our deep learning solver generates very smooth results with no such artifacts.

%Moreover, and its absolute energy value doesn't represent the quality of visual results

Given each specific $L_p$ distribution map in each iteration, the traditional numerical solvers still tend to ``overfit'' that unique distribution map for its optimal results, which accordingly results in some spurious edges that separate these different regularizations just like the case of half $L_2$ half $L_{0.8}$ norm. In contrast, the disadvantage of the deep learning solver exposed in Section~\ref{sec:fixedlp} becomes an advantage in the presence of a dynamically changed $L_p$ norm distribution. \emph{Benefited from the large corpus of training data, the deep learning solver incorporates the learned implicit combination of $L_2$ and $L_{0.8}$ norm into the deep network and reflects such combination, instead of a fixed regularizer, into each pixel of the smoothed images}. It is able to generates more visually pleasing results, as shown in both Figure~\ref{figure:comp_network} and ~\ref{figure:loss_images}.

Therefore, we argue that what matters to solve the proposed objective function and obtain better smoothing result is the joint optimization over large corpus of images, instead of any particular image. In this specific problem, the deep learning solver plays a critical role. Note that although many empirical experiments have been conducted above, rigorous theoretical analysis is still lacking. Understanding and explaining deep neural networks are still open problems for the follow-up research. 

\section{Experimental Results}\label{sec:experimentalanalysis}

In this section, we first conduct some ablation study to analyze the influence of the parameters and network structures to the results. Afterwards, we compare our results with previous methods in both visual quality and running time efficiency.

\setlength{\tabcolsep}{1pt}
\begin{figure*}[htp]
\begin{center}
\begin{tabular}{c@{\hskip 1mm} | @{\hskip 1mm}cccc}
\includegraphics[width=3.5cm,trim={0 0 0 0},clip]{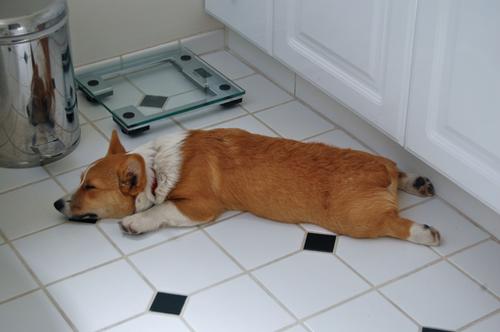}
&\includegraphics[width=3.5cm,trim={0 0 0 0},clip]{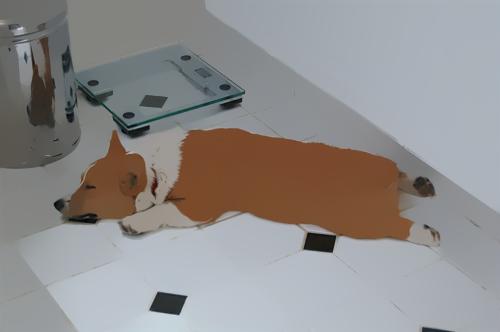}
&\includegraphics[width=3.5cm,trim={0 0 0 0},clip]{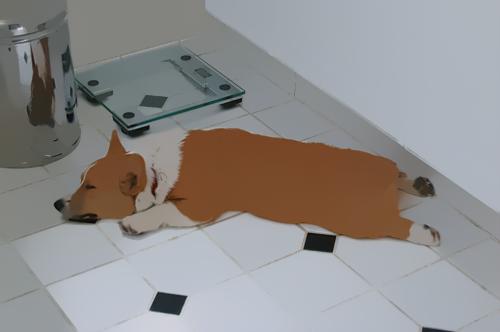}
&\includegraphics[width=3.5cm,trim={0 0 0 0},clip]{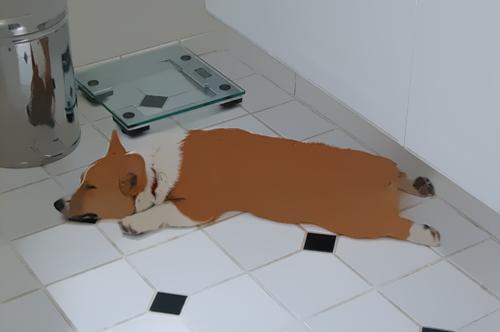}
&\includegraphics[width=3.5cm,trim={0 0 0 0},clip]{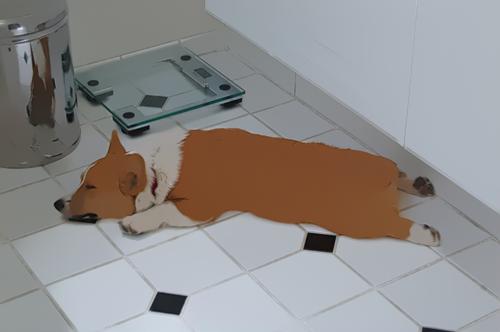}
\vspace{-0.6mm}\\

%\vspace{1mm}
Input & $\lambda_f = 1, \lambda_e = 0.01$ & $\lambda_f = 1, \lambda_e = 0.025$ & $\lambda_f = 1, \lambda_e = 0.1$ & $\lambda_f = 1, \lambda_e = 0.2$
\\

\\[-0.9em]

\includegraphics[width=3.5cm,trim={0 0 0 0},clip]{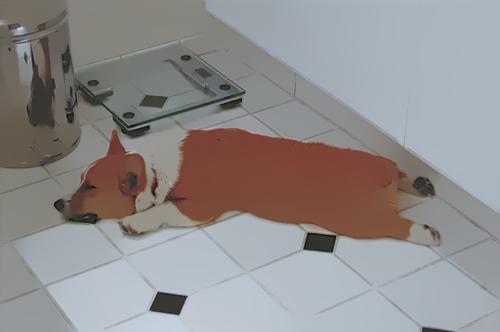}
&\includegraphics[width=3.5cm,trim={0 0 0 0},clip]{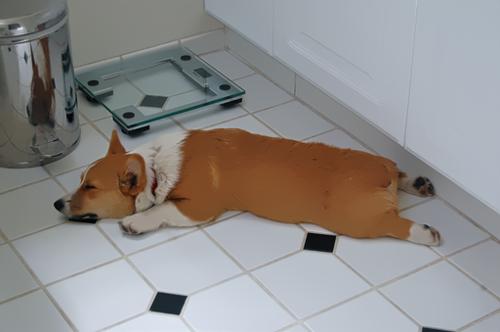}
&\includegraphics[width=3.5cm,trim={0 0 0 0},clip]{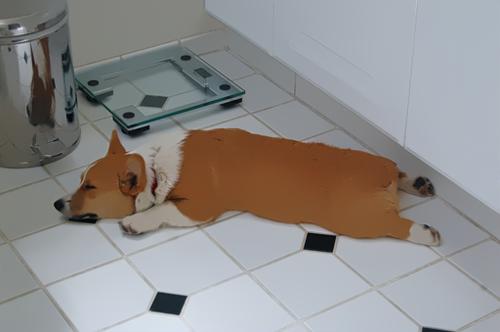}
&\includegraphics[width=3.5cm,trim={0 0 0 0},clip]{smooth-63-input-predict_standard_edgePreserving_L2_weight5_window3_threshold_20_10_window10_L08_w1_epoch30.jpg}
&\includegraphics[width=3.5cm,trim={0 0 0 0},clip]{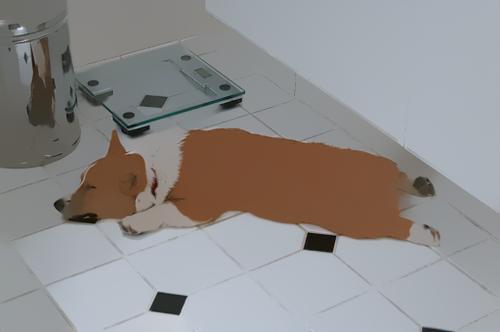}
\vspace{-0.6mm}\\

w/o Residual Learning & $\lambda_f = 0.1, \lambda_e = 0.1$ & $\lambda_f = 0.25, \lambda_e = 0.1$ & $\lambda_f = 1, \lambda_e = 0.1$ & $\lambda_f = 2, \lambda_e = 0.1$
\\

\end{tabular}
\end{center}
\vspace{-2mm}
\caption{Effect of tuning parameters (Right) and ablation study for our residual learning (Left). The first row shows that with larger $\lambda_e$, we are able to maintain some important image structures. The second row shows that via adjusting $\lambda_f$, we can gradually change the smoothness strength. As can be seen on the left, the result predicted by the model without residual connection exhibits some noticeable color attenuation problems, which does not exist in our results with residual image learning. Photo courtesy of Flickr user Rachel Hinman.}
\label{figure:parameter}
\vspace{1mm}
\end{figure*}

\setlength{\tabcolsep}{5pt}
\renewcommand{\arraystretch}{1.0}
\begin{table*}[t]
\vspace{-2mm}
\caption{Running time comparison between different methods (in seconds). Our method is significantly faster than the traditional methods (SGF~\cite{zhang2015segment}, BLF~\cite{BLF98}, RGF~\cite{RGF}, Tree Filter~\cite{tree}, $L_0$~\cite{L0smooth}, RTV~\cite{RTV}, WLS~\cite{WLS}, SDF~\cite{ham2015robust} and $L_1$~\cite{L1smooth}), especially those based on optimization ($L_0$, RTV, WLS, SDF, $L_1$). It is also much faster than the recent methods~\cite{xu2015,liu2016,fan2017generic} that train convolutional neural networks (CNN) as a regression tool to approximate traditional smoothing approaches. Due to lack of current GPU implementation of most traditional methods, their running time is evaluated under a modern CPU. The deep learning based methods are evaluated on GPU, meanwhile we also report the CPU time of our network structure with efficient multi-core implementation.} \label{table:time}
\vspace{-4mm}
	\begin{center}
		\begin{tabular}{ r | c c c c c c c c c | c c c | c c}
            %\toprule[0.08em]
            \hline
            \multirow{2}{*}{\small{ }} &\multicolumn{9}{ c |}{\small{Traditional smoothing algorithms}} & \multicolumn{3}{ c |}{\small{Approximation CNN}} & \multicolumn{2}{ c }{\small{Ours}} \\
            %\midrule
            \cline{2-15}
			\small{ } & \small{SGF} & \small{BLF} & \small{RGF} & \small{Tree} & \small{$L_0$} & \small{RTV} & \small{WLS} & \small{SDF} & \small{$L_1$} & \small{Xu} & \small{Liu} & \small{Fan} & GPU & CPU\\
			%\midrule
			\hline
%			{QVGA (320$\times$240)} &0.05 &0.03 &0.22 & 0.05 &0.17 &0.41 &0.70 &4.99 &32.18 &0.23 &0.07 & 0.008 &0.003 & 2.75\\
%			{VGA (640$\times$480)} &0.15 &0.12 &0.73 &0.42 &0.66 &1.80 &3.34 &19.19 &212.07 &0.76 &0.14 & 0.009 & 0.004 & 9.61\\
%			{720p (1280$\times$720)} &0.25 &0.34 &1.87 & 2.08 &2.43 &5.74 &13.26 &66.14 &904.36 &2.16 &0.33 & 0.010 & 0.005 & 25.62\\
			{QVGA (320$\times$240)} &0.05 &0.03 &0.22 & 0.05 &0.17 &0.41 &0.70 &4.99 &32.18 &0.23 &0.07 & 0.008 &0.003 & 0.010\\
			{VGA (640$\times$480)} &0.15 &0.12 &0.73 &0.42 &0.66 &1.80 &3.34 &19.19 &212.07 &0.76 &0.14 & 0.009 & 0.004 & 0.011\\
			{720p (1280$\times$720)} &0.25 &0.34 &1.87 & 2.08 &2.43 &5.74 &13.26 &66.14 &904.36 &2.16 &0.33 & 0.010 & 0.005 & 0.012\\
			\hline
			%\bottomrule
		\end{tabular}
	\end{center}
\end{table*}

\subsection{Ablation Study}

\paragraph{Effect of parameter control}
%As there are too many parameters that can be tweaked to influence the performance of our algorithm, it's impossible to demonstrate the effects of there parameters one by one in the paper. Thus we choose to experiment with two most important ones $\lambda_f$ and $\lambda_e$, as all the other parameters can be incorporated in the energy functions weighted by them.

In out method, the main parameters to tune are $\lambda_f$ and $\lambda_e$. Here we analyze the results of our network trained under different settings of these two parameters, and such a group of visual results are shown in Figure~\ref{figure:parameter}. As can be seen from the first row, altering weight of edge-preserving term $\lambda_e$ influences the image structures. From the second row, tweaking the weight for flattening term $\lambda_f$ controls the smoothness of predicted images. While enhancing the smoothness with a large $\lambda_f$, we observe gradually destructed structures, \emph{e.g.} the ground tiles. 
%So it controls a trade-off between edge preservation and detail elimination.

\paragraph{Effect of residual image learning}
%We discuss about a few critical ingredients of the network that may influence the performance in this section.

We analyze the importance of residual image learning by comparing the results with and without this component in our network. As shown in Figure~\ref{figure:parameter}, the smooth image generated without residual learning contains some obvious color degradation issues. It appears more orange compared to the raw input image. In contrast, the smooth images predicted with our complete network structure with residual image learning dose not have this issue, as shown in Figure~\ref{figure:parameter}. 
For the image smoothing task, the input and output image should be highly correlated. However, it can be difficult to maintain well the color information in the image after many convolution operations in a deep neural network like ours. Thus we propose to learn the residual image and combine it with input image to resolve this issue.

%In addition, the network initialization \cite{delve} and residual block \cite{he2016deep} in our framework are very common techniques that have been adopted by many previous work. We experiment with other alternatives, including bigger batch size, but didn't observed obvious improvement.

\setlength{\tabcolsep}{1pt}
\begin{figure*}[t]
\begin{center}
\begin{tabular}{c ccc ccc c}
\vspace{0.5mm}
&  \multicolumn{3}{ c }{Image abstraction}  & \multicolumn{3}{ c }{Pencil drawing} & \\

\includegraphics[width=2.1cm,trim={0 1.5cm 0 0},clip]{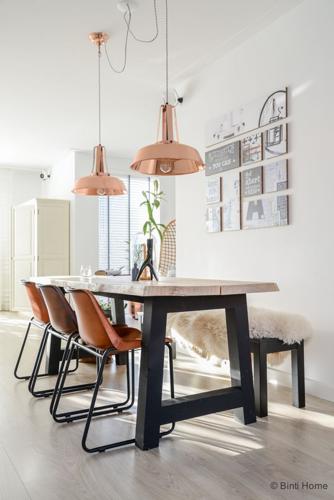}
&\includegraphics[width=2.1cm,trim={0 1.5cm 0 0},clip]{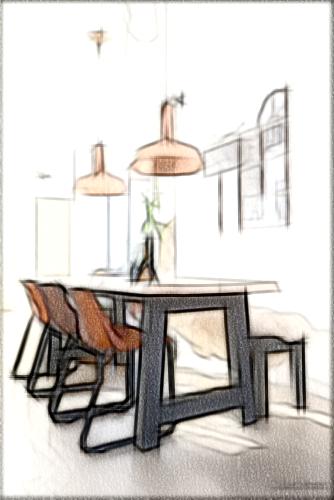}
&\includegraphics[width=2.1cm,trim={0 1.5cm 0 0},clip]{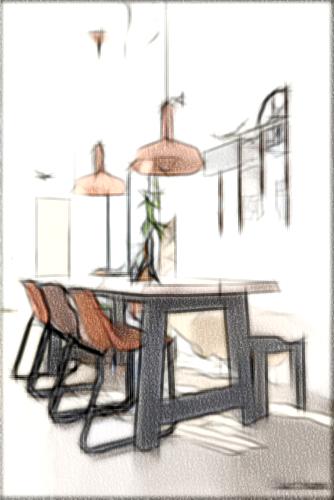}
&\includegraphics[width=2.1cm,trim={0 1.5cm 0 0},clip]{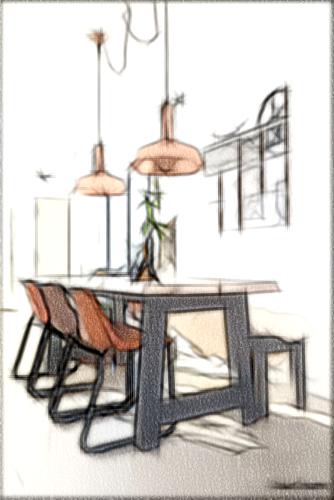}
&\includegraphics[width=2.1cm,trim={0 1.5cm 0 0},clip]{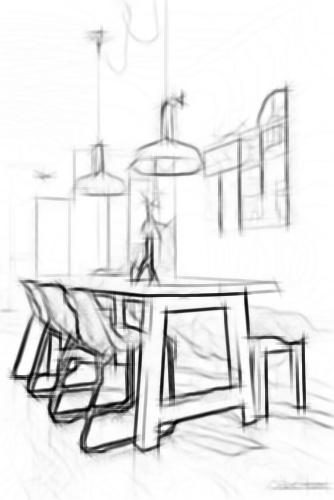}
&\includegraphics[width=2.1cm,trim={0 1.5cm 0 0},clip]{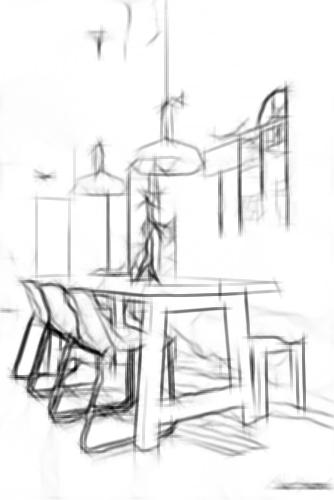}
&\includegraphics[width=2.1cm,trim={0 1.5cm 0 0},clip]{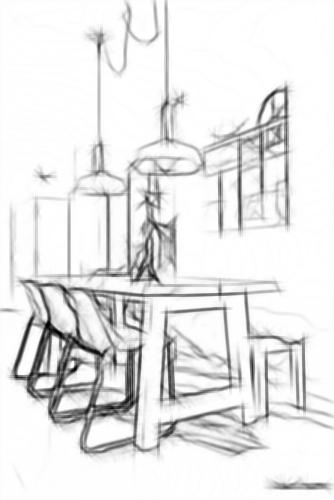}
&\includegraphics[width=2.1cm,trim={0 1.5cm 0 0},clip]{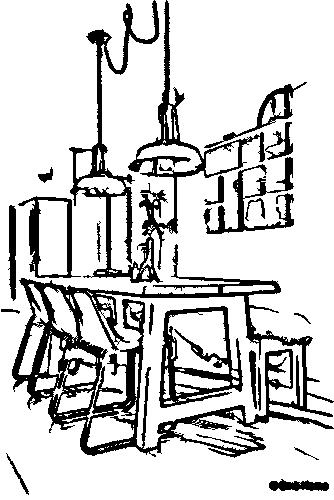}
\\

\includegraphics[width=2.1cm,trim={2cm 4cm 0 0},clip]{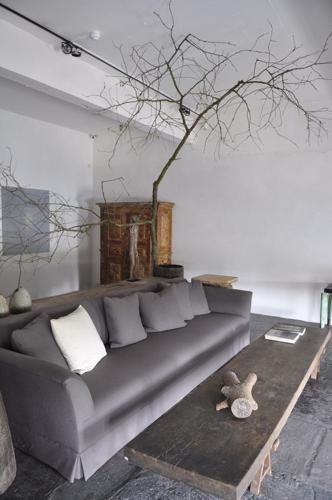}
&\includegraphics[width=2.1cm,trim={2cm 4cm 0 0},clip]{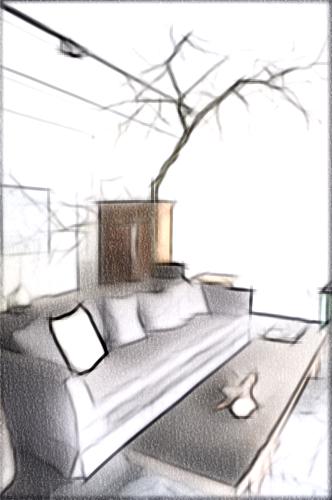}
&\includegraphics[width=2.1cm,trim={2cm 4cm 0 0},clip]{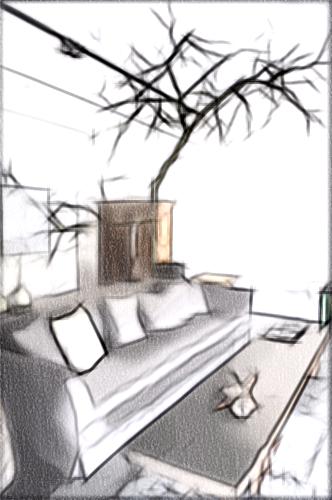}
&\includegraphics[width=2.1cm,trim={2cm 4cm 0 0},clip]{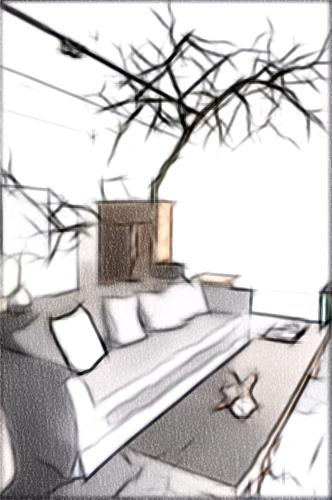}
&\includegraphics[width=2.1cm,trim={2cm 4cm 0 0},clip]{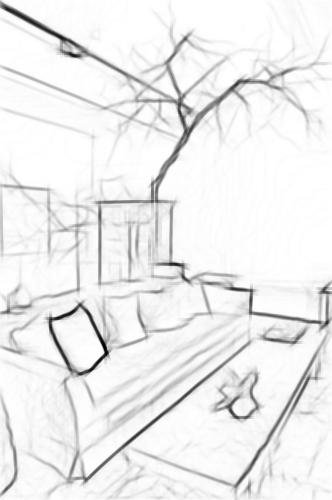}
&\includegraphics[width=2.1cm,trim={2cm 4cm 0 0},clip]{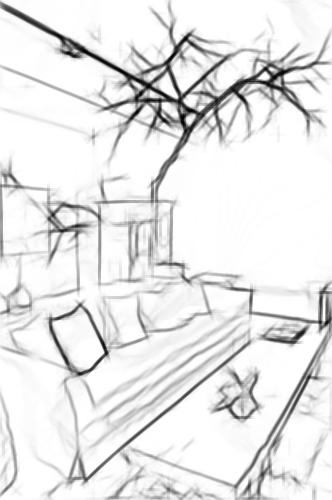}
&\includegraphics[width=2.1cm,trim={2cm 4cm 0 0},clip]{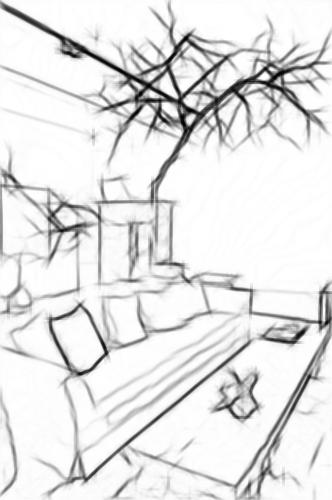}
&\includegraphics[width=2.1cm,trim={2cm 4cm 0 0},clip]{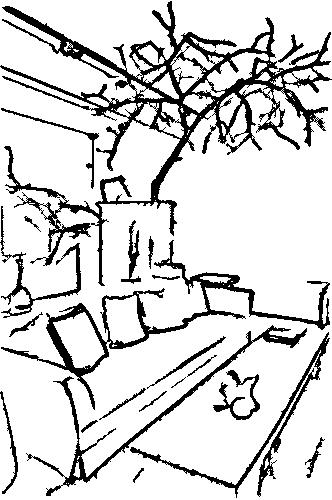}
\\

Input & WLS & $L_0$ & Ours & WLS & $L_0$ & Ours & $B$
\\

\end{tabular}
\end{center}
\vspace{-3mm}
\caption{Comparison of the image abstraction (Column 2-4) and pencil sketching (Column 5-7) results between our method and previous image smoothing methods WLS~\cite{WLS} and~$L_0$ \cite{L0smooth}. With our learned edge-preserving image smoother, the slender but important image structures such as the lamp wires and twigs are well preserved, rendering the stylized images more visually pleasing. Moreover, we also demonstrate the binary edge map $B$ detected by our heuristic detection method, which shows consistent image structure with our style image. Note that binary edge maps are only used in the objective function for training; they are not used in the test stage and are presented here only for comparison purpose.}
\label{figure:stylization}
\vspace{1mm}
\end{figure*}

\subsection{Comparison with Previous Methods}

We compare the proposed method with previous ones in terms of the smoothing effect and speed. More comparisons on different applications can be found in Section~\ref{sec:applications}.

\paragraph{Smoothing effect comparison} Figure~\ref{figure:comp_others} compares the smoothing results of our method and ten existing methods: \cite{zhang2015segment} (SGF), \cite{ham2015robust} (SDF), \cite{L1smooth} ($L_1$), \cite{cho2014bilateral} (BTLF), \cite{FGS} (FGS), \cite{RGF} (RGF), \cite{RTV} (RTV), \cite{L0smooth} ($L_0$), \cite{WLS} (WLS) and \cite{BLF98} (BLF). Note that these algorithms may be designed for different applications thus their goals may be slightly different. Compared to these methods under this difficult example, our method produced outstanding edge-preserving flattening result: it not only achieved pleasing flattening effects for regions of low amplitude (\emph{e.g.} the sea), but also well preserved the high-contrast structures, especially the thin but salient edges (\emph{e.g.} the ropes of the paraglider).

To our knowledge, there is no benchmark or dataset to quantitatively evaluate the performance of image smoothing algorithms. Visual perception is still the principal way for evaluation. To demonstrate the robustness of our method and its good performance for natural images with vastly different contents and capturing conditions, we present the visual results on over 100 images without any parameter tweaking for any particular images in the supplemental material.

%Thus to avoid the suspect of cherry-picking good results to show in the paper, we demonstrate the success of our algorithm on over 100 images without any parameter tweaking for any particular images and put these results in the supplemental material.

\paragraph{Running time comparison} Table \ref{table:time} compares the running time of our method and some previous methods, including both traditional image smoothing algorithms \cite{zhang2015segment,BLF98,RGF,tree,L0smooth,RTV,WLS,ham2015robust,L1smooth} and some recent methods \cite{xu2015,liu2016,fan2017generic} that apply neural networks to approximate the results of the existing smoothing algorithms. Traditional image smoothing methods are based on either filtering techniques~\cite{ham2015robust,BLF98,RGF,tree} or mathematical optimization~\cite{L1smooth,L0smooth,WLS,RTV,ham2015robust}. While the latter category draws much attention in recent years and often produces quality results, the optimization procedure (\emph{e.g.} solving large-scale linear systems iteratively) can be very time-consuming. For example, the state-of-the-art method of \cite{L1smooth} takes about 15 minutes to process a 1280$\times$720 image.
Compared to these methods, ours runs significantly faster. It takes only a few milliseconds for a 1280$\times$720 image at the aid of GPU. However, even on CPU, with efficient parallel implementation of our network structure\footnote{Our CPU version is implemented in MXNet facilitated with the NNPACK module.}, it still runs in at most tens of milliseconds, facilitating real-time applications.

Compared to the neural network approximators~\cite{xu2015,liu2016,fan2017generic}\footnote{The running time of \cite{fan2017generic} reported in their paper includes both generating images of particular sizes (per Table~\ref{table:time}) and running the network. We excluded the former for a fair comparison, thus their numbers are lower than reported.}, our method not only generates novel, unique smoothing effects that allow better results in various applications (see Section~\ref{sec:applications}), but also has a faster running speed. Note except for running a neural network, \cite{xu2015} also employs a post-processed optimization step and \cite{liu2016} leverages a recursive 1D filter, both of which slows down their whole algorithm. 

\section{Applications}\label{sec:applications}

In this section, we demonstrate the effectiveness and flexibility of our image smoothing algorithm with a range of different applications including image abstraction, pencil sketching, detail magnification, texture removal and content-aware image manipulation. The tailored methods for these different applications mainly differ in the guidance edge maps used in training the network: the former three applications use the local gradient based edge map (per Equation~\ref{eq:Edge_forward}) similar to the previous experiments, while the latter two modify them to achieve particular effects. For all the applications we use the images in the PASCAL VOC dataset~\cite{everingham2010pascal} for training.

\setlength{\tabcolsep}{1pt}
\begin{figure*}[t]
\begin{center}
\begin{tabular}{cccccc}

%\includegraphics[width=2.8cm]{flower2-33-input_vis.jpg}
%&\includegraphics[width=2.8cm]{flower2-33-input-label-BLF_vis.jpg}
%&\includegraphics[width=2.8cm]{flower2-33-input-label-WLS_vis.jpg}
%&\includegraphics[width=2.8cm]{flower2-33-input-label-L0smooth_vis.jpg}
%&\includegraphics[width=2.8cm]{flower2-33-input-label-FGS_vis.jpg}
%%&\includegraphics[width=2.8cm]{flower2-33-input-predict_standard_enhance_L2_weight9_window0_threshold0_epoch30.ng}
%&\includegraphics[width=2.8cm]{flower2-33-input-predict_standard_enhance_L2_weight15_window0_threshold0_window10_L08_w1_epoch30_vis.jpg}
%\\
%Input image & BLF smooth & WLS smooth & $L_0$ smooth & FGS smooth & Ours smooth
%\\
%
%\includegraphics[width=2.8cm]{flower2-33-input-LLF_vis.jpg}
%&\includegraphics[width=2.8cm]{flower2-33-input-label-BLF-detail_vis.jpg}
%&\includegraphics[width=2.8cm]{flower2-33-input-label-WLS-detail_vis.jpg}
%&\includegraphics[width=2.8cm]{flower2-33-input-label-L0smooth-detail_vis.jpg}
%&\includegraphics[width=2.8cm]{flower2-33-input-label-FGS-detail_vis.jpg}
%%&\includegraphics[width=2.8cm]{flower2-33-input-predict_standard_enhance_L2_weight9_window0_threshold0_epoch30-detail.jpg}
%&\includegraphics[width=2.8cm]{flower2-33-input-predict_standard_enhance_L2_weight15_window0_threshold0_window10_L08_w1_epoch30-detail_vis.jpg}
%\\
%LLF enhance & BLF enhance & WLS enhance & $L_0$ enhance & FGS enhance & Ours enhance
%\\

\includegraphics[width=2.92cm,trim={0.6cm 0.6cm 0.6cm 0.6cm},clip]{sig_v2_input_76.jpg}
&\includegraphics[width=2.92cm,trim={0.6cm 0.6cm 0.6cm 0.6cm},clip]{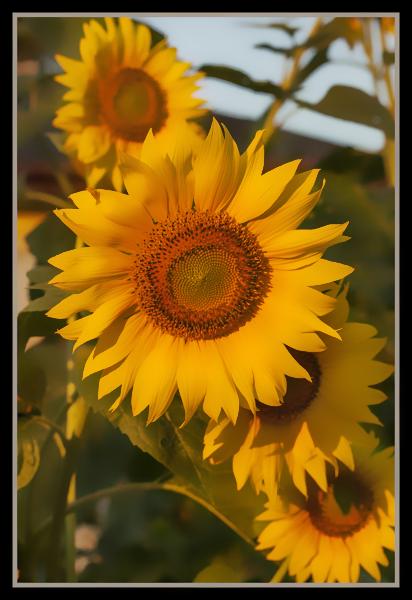}
&\includegraphics[width=2.92cm,trim={0.6cm 0.6cm 0.6cm 0.6cm},clip]{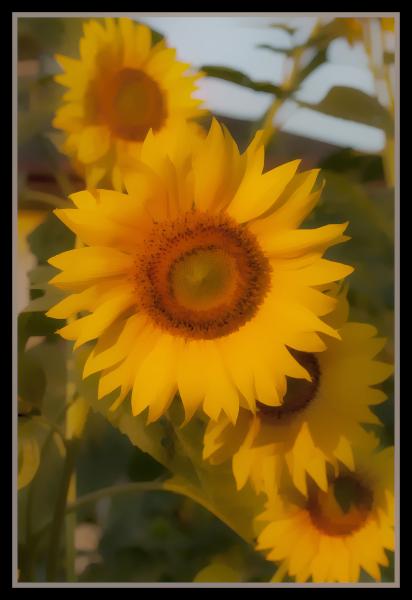}
&\includegraphics[width=2.92cm,trim={0.6cm 0.6cm 0.6cm 0.6cm},clip]{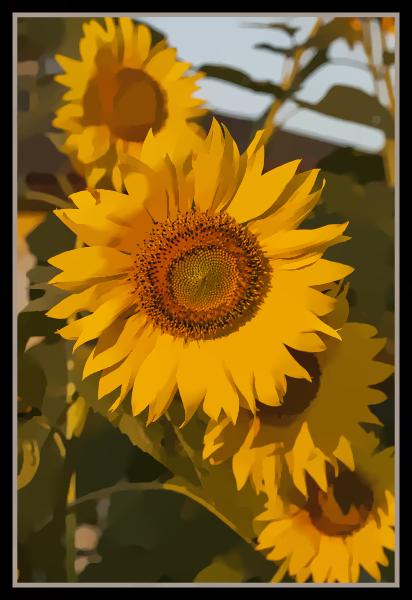}
&\includegraphics[width=2.92cm,trim={0.6cm 0.6cm 0.6cm 0.6cm},clip]{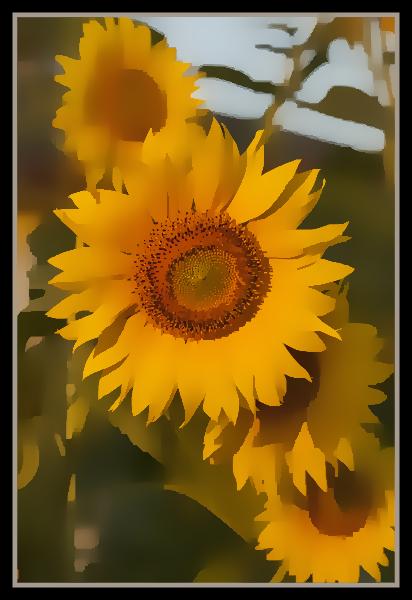}
%&\includegraphics[width=2.92cm]{sig_v2_input_76-predict_standard_enhance_L2_weight9_window0_threshold0_epoch30.ng}
&\includegraphics[width=2.92cm,trim={0.6cm 0.6cm 0.6cm 0.6cm},clip]{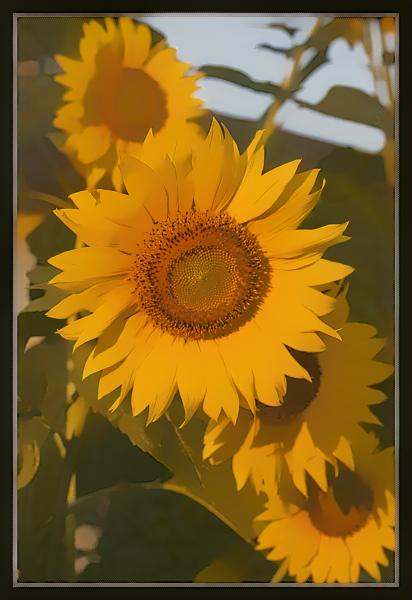}
\vspace{-1pt}\\
Input image & BLF smooth & WLS smooth & $L_0$ smooth & FGS smooth & Ours smooth
\vspace{3pt}\\

\includegraphics[width=2.92cm,trim={0.6cm 0.6cm 0.6cm 0.6cm},clip]{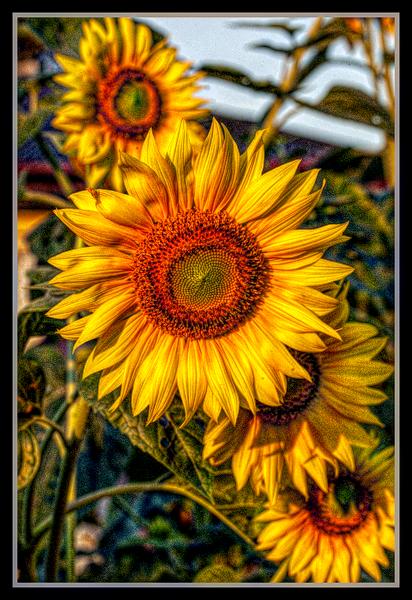}
&\includegraphics[width=2.92cm,trim={0.6cm 0.6cm 0.6cm 0.6cm},clip]{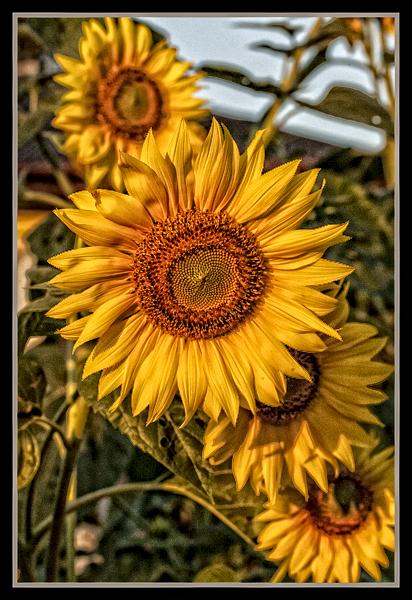}
&\includegraphics[width=2.92cm,trim={0.6cm 0.6cm 0.6cm 0.6cm},clip]{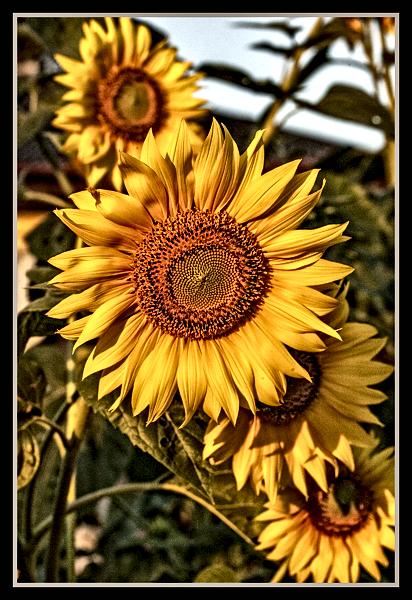}
&\includegraphics[width=2.92cm,trim={0.6cm 0.6cm 0.6cm 0.6cm},clip]{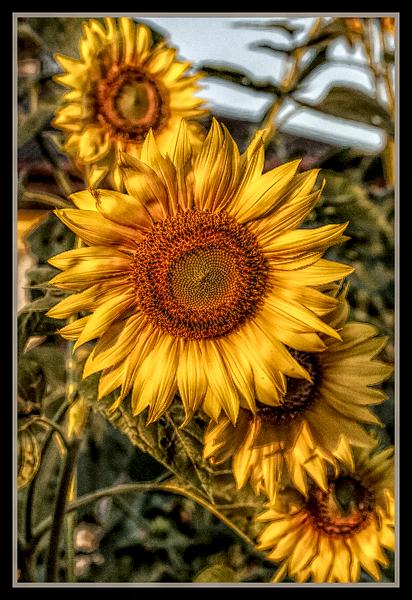}
&\includegraphics[width=2.92cm,trim={0.6cm 0.6cm 0.6cm 0.6cm},clip]{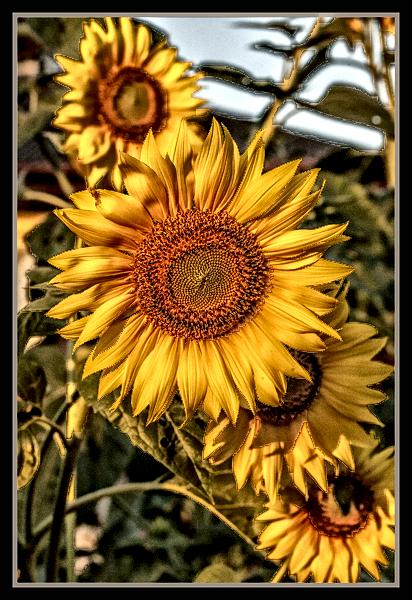}
%&\includegraphics[width=2.92cm]{sig_v2_input_76-predict_standard_enhance_L2_weight9_window0_threshold0_epoch30-detail.jpg}
&\includegraphics[width=2.92cm,trim={0.6cm 0.6cm 0.6cm 0.6cm},clip]{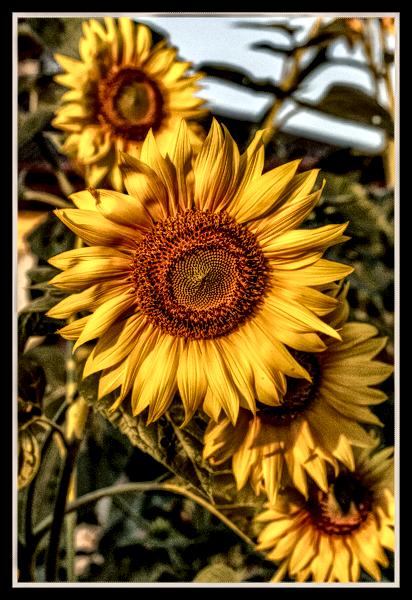}
\vspace{-1pt}\\
LLF enhance & BLF enhance & WLS enhance & $L_0$ enhance & FGS enhance & Ours enhance
\end{tabular}
\end{center}
\vspace{-2mm}
\caption{Detail magnification results of our method compared with previous image smoothing algorithms LLF ~\cite{paris2011local}, BLF~\cite{BLF98}, WLS~\cite{WLS}, $L_0$~\cite{L0smooth} and FGS~\cite{FGS}. In this example, the top row shows the smooth base layers obtained via image smoothing, while the bottom one demonstrates the enhancement results. As can be seen, our algorithm does not over-sharpen the image structures in the smooth image and achieves visually pleasing detail exaggeration effects. \textbf{Zoom in} to see the details. Photo courtesy of Flickr user Sheba\_Also.}
\label{figure:detail}
\vspace{1mm}
\end{figure*}

\subsection{Image Abstraction and Pencil Sketching} \label{abstraction}
Edge-preserving image smoothing can be used to stylize imageries. For example, \cite{winnemoller2006real} proposed to abstract imagery by simplifying the low-amplitude details and increasing the contrast of visually important structures with difference-of-Gaussian edges. Following previous works of \cite{WLS,L0smooth}, we replace the iterative bilateral filter in \cite{winnemoller2006real} with our learned edge-preserving image smoother, and decorate the extracted edges with random sketches in different directions to generate pencil drawing pictures. Furthermore, the smoothed images are combined with the pencil drawing picture to generate an abstract look \cite{lu2012combining}.

Figure~\ref{figure:stylization} presents the image abstraction and pencil sketching results of different methods on two examples. Our method clearly excelled at preserving important image structures, thanks to the proposed energy function which has an explicit edge-preserving constraint. For example, the lamp wires in the first example can be clearly seen in our smoothing results, while they are not well preserved by other methods. Note that these structures are semantically meaningful, without which the images look strange. The abstraction and pencil sketching results of our method are clearly more satisfactory. In the second example, the tree branches are small and thin, but are still visually prominent in this image. Our method well kept the tree structure, while \cite{WLS} only preserved the limbs and \cite{WLS} broke some thin branches into pieces.

Note to further overcome the over-sharpened effects, we expand the image region regularized by $L_{p^{large}}$ norm to its surrounding 7$\times$7 pixel neighbourhood.

\setlength{\tabcolsep}{1pt}
\begin{figure*}[t]
\begin{center}
\begin{tabular}{ccccccc}

%\vspace{-0.5mm}
%\includegraphics[width=2.4cm]{texture_smooth_e1-input.jpg}
%&\includegraphics[width=2.4cm]{texture_smooth_e1-input-RTV.jpg}
%&\includegraphics[width=2.4cm]{texture_smooth_e1-input-RGF.jpg}
%&\includegraphics[width=2.4cm]{texture_smooth_e1-input-BTLF.jpg}
%&\includegraphics[width=2.4cm]{texture_smooth_e1-input-SGF.jpg}
%&\includegraphics[width=2.4cm]{texture_smooth_e1-input-SDF.jpg}
%&\includegraphics[width=2.4cm]{texture_smooth_e1-input-predict_standard_texture_L2_weight20_window0_threshold_20_10_window2_L08_w1_epoch30.jpg}
%%&\includegraphics[width=2.4cm]{texture_smooth_e1-input-edge-texture.jpg}
%\\

%\vspace{-0.5mm}
%\includegraphics[width=2.4cm]{RTV-192.jpg}
%&\includegraphics[width=2.4cm]{RTV-192-RTV.jpg}
%&\includegraphics[width=2.4cm]{RTV-192-RGF.jpg}
%&\includegraphics[width=2.4cm]{RTV-192-BTLF.jpg}
%&\includegraphics[width=2.4cm]{RTV-192-SGF-4.jpg}
%&\includegraphics[width=2.4cm]{RTV-192-SDF-100.jpg}
%&\includegraphics[width=2.4cm]{RTV-192-predict_standard_texture_L2_weight20_window0_threshold_20_10_window2_L08_w1_epoch30.jpg}
%%&\includegraphics[width=2.4cm]{RTV-192-edge-texture.jpg}
%\\

\includegraphics[width=2.49cm]{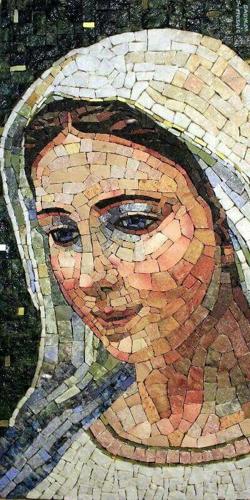}
&\includegraphics[width=2.49cm]{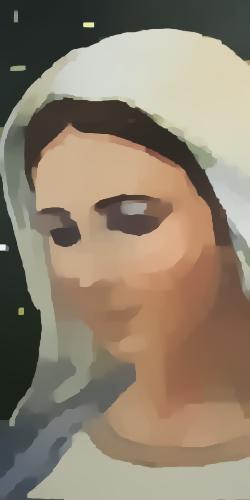}
&\includegraphics[width=2.49cm]{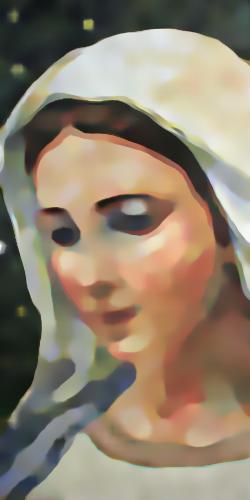}
&\includegraphics[width=2.49cm]{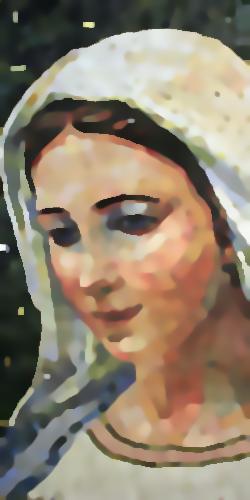}
&\includegraphics[width=2.49cm]{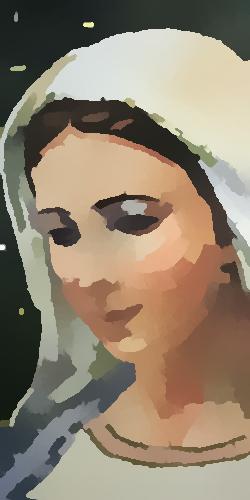}
&\includegraphics[width=2.49cm]{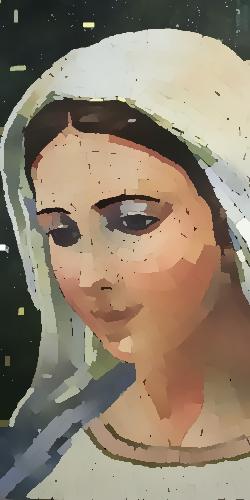}
&\includegraphics[width=2.49cm]{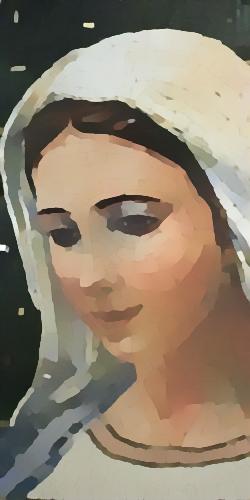}
%&\includegraphics[width=2.49cm]{texture_smooth_e1-input-edge-texture.jpg}
\vspace{-0.3mm}\\

\includegraphics[width=2.49cm]{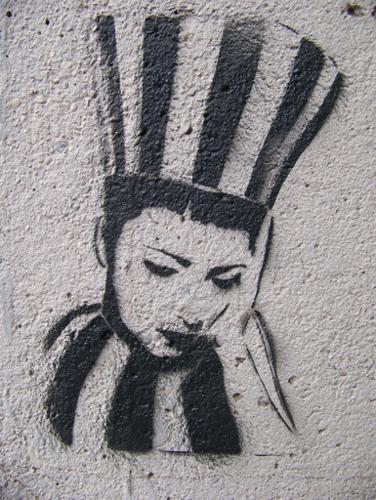}
&\includegraphics[width=2.49cm]{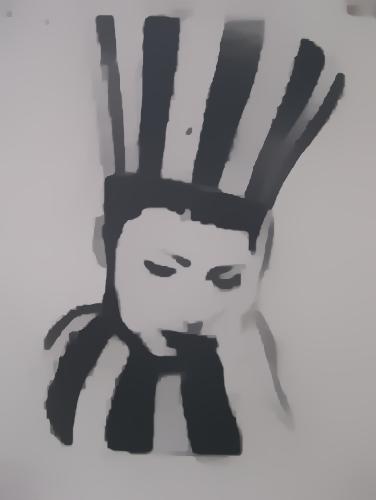}
&\includegraphics[width=2.49cm]{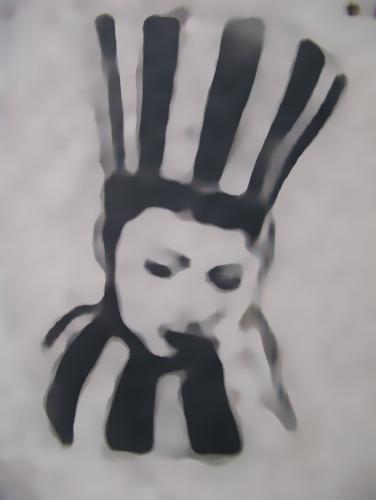}
&\includegraphics[width=2.49cm]{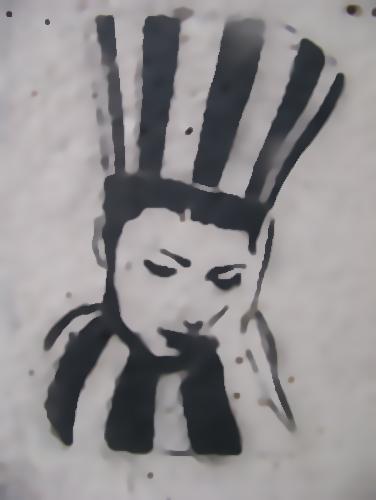}
&\includegraphics[width=2.49cm]{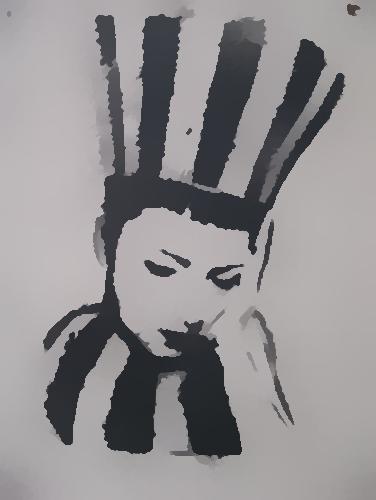}
&\includegraphics[width=2.49cm]{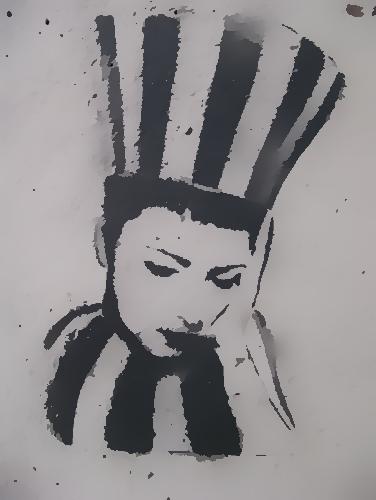}
&\includegraphics[width=2.49cm]{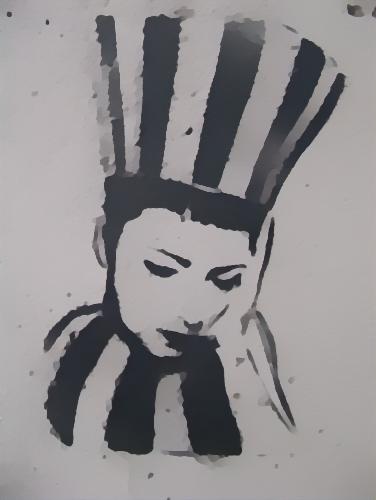}
%&\includegraphics[width=2.49cm]{RTV-192-edge-texture.jpg}
\vspace{-0.2mm}\\
Input & RTV & RGF & BTLF & SGF & SDF & Ours
\\
\end{tabular}
\end{center}
\vspace{-2mm}
\caption{Texture removal results of our method compared with state-of-the-art methods that address the texture removal problem: RTV~\cite{RTV}, RGF~\cite{RGF}, BTLF~\cite{cho2014bilateral}, SGF~\cite{zhang2015segment} and SDF~\cite{ham2015robust}. Our method can effectively remove the texture patterns meanwhile well preserve the primary image structures. Photo courtesy of Emilie Baudrais and \cite{RTV}.} %Marie-Lan Nguyen
\label{figure:texture}
%\vspace{2mm}
\end{figure*}

\subsection{Detail Magnification} \label{detail}

The effect of image detail magnification can be achieved by superposing a smooth base layer and an enhanced detail layer, the latter of which can be obtained by image smoothing algorithms. After extracting the smooth layer, the detail layer can be obtained as the difference between the original image and the smooth layer, which is then enhanced and added back to the smooth layer to generate the final result. An ideal smoothing algorithm for this task should neither blur nor over-sharpen the salient image structures~\cite{WLS}, as either operation can lead to ``ringing'' artifacts in the residual image, resulting in halo or gradient reversals in the detail-enhanced images. Developing such a smooth filter is challenging as it is difficult to determine the edges to preserve and diminish while avoiding to both over sharpen and smooth these edges.

Figure~\ref{figure:detail} presents the results on such example obtained by our method and previous ones~\cite{L0smooth,WLS,BLF98,FGS,paris2011local}. It can be observed that in the smoothed images the methods of \cite{L0smooth,BLF98,FGS} sharpened some edges that are blurry in the original images due to out of focus. As a result, conspicuous gradient reversal artifacts can be observed clearly on the top of enhanced images. In contrast, \cite{WLS,paris2011local} and our method produce better results without noticeable artifacts. Note that the method of \cite{WLS} applied $L_2$ regularizer over the entire image in their smoothness term to perfectly avoid over-sharpening the structures. In our approach, the edge-preserving term enforces a strong similarity between the major image structure of input and output images via minimizing their quadratic differences, preventing the edges from being significantly blurred or excessively sharpened. Moreover, the $L_2$ norm is also partially applied to the potentially over-sharpened regions to better avoid gradient reversal artifacts in the exaggerated image.
As such, high-quality detail magnification results can be obtained as shown in Figure~\ref{figure:detail}.

Note the gradient reversal artifacts are very likely to happen even if the smooth image is only slightly over-sharpened by either numerical analysis or visual perception. And such a case is almost unavoidable as for the edge-preserving filters that applies strong regularization, since it always tends to over-sharpen the image more or less. Therefore, we do not argue for the perfect detail exaggeration results, but we are able to outperform most previous algorithms that pursue strong smoothing effects~\cite{BLF98,FGS,L0smooth} with only little effort of tweaking our objective function.

To avoid the gradient reversal effects that are usually caused by over-sharpening the smooth layer, we increase the $p^{large}$ balance weight ($\alpha = 15$), and release the constraint on $p$ selection ($c_1 = +\infty, c_2 = 0$).

\setlength{\tabcolsep}{1pt}
\begin{figure}[t]
\begin{center}
\begin{tabular}{ccc}

\includegraphics[width=2.8cm,clip]{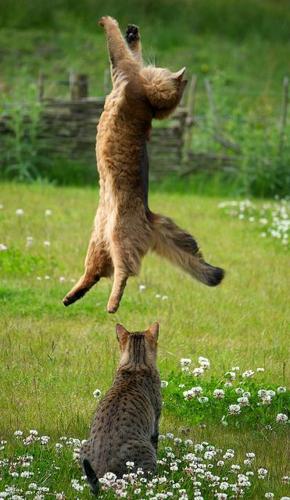}
&\includegraphics[width=2.8cm,clip]{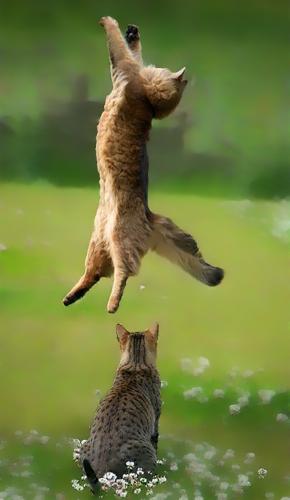}
&\includegraphics[width=2.8cm,clip]{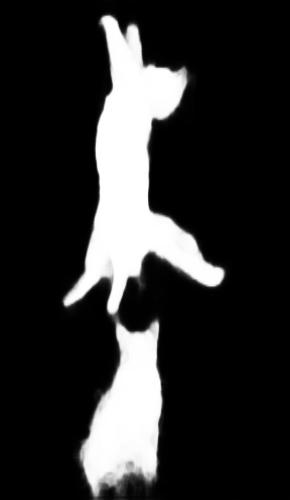}
\vspace{-1pt}\\

%\includegraphics[width=2.8cm]{blur10-500-7-input.jpg}
%&\includegraphics[width=2.8cm]{blur10-500-7-input-predict_standard_blur_L2_weight5_window0_threshold_20_10_window2_L08_w0_epoch30.jpg}
%&\includegraphics[width=2.8cm]{blur10-500-7-input-saliency.jpg}
%\\

\small{Input} & \small{Background smoothed} & \small{Saliency Map}
\vspace{3pt}\\

\includegraphics[width=2.8cm]{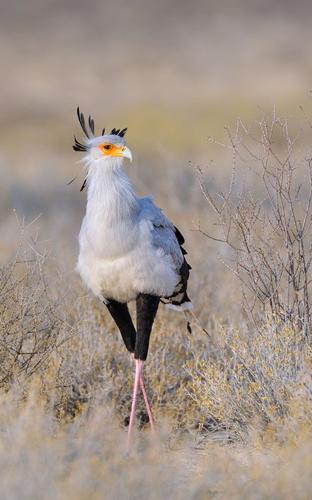}
&\includegraphics[width=2.8cm]{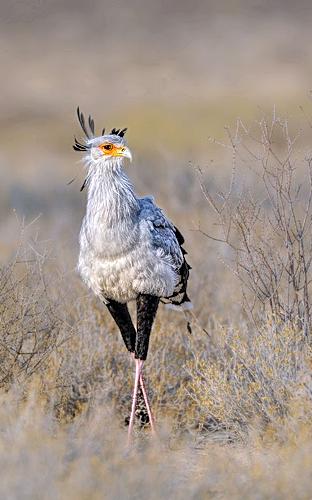}
&\includegraphics[width=2.8cm]{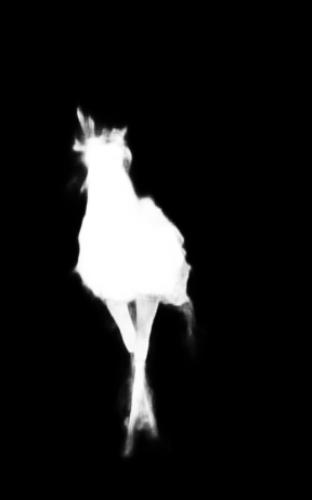}
\vspace{-1pt}\\

%\includegraphics[width=2.8cm]{blur-previous-5.jpg}
%&\includegraphics[width=2.8cm]{blur-previous-5-predict30-detail.jpg}
%&\includegraphics[width=2.8cm]{blur-previous-5-saliency.jpg}
%\\

\small{Input} & \small{Foreground enhanced} & \small{Saliency Map}
\\

%\includegraphics[width=2.8cm,trim={0 2cm 0 0},clip]{blur11-500-3-input.jpg}
%&\includegraphics[width=2.8cm,trim={0 2cm 0 0},clip]{blur11-500-3-input-predict_standard_blur_L2_weight5_window0_threshold_20_10_window2_L08_w0_epoch30.jpg}
%&\includegraphics[width=2.8cm,trim={0 2cm 0 0},clip]{blur11-500-3-input-predict_standard_saliency_sharpen_L2_weight5_window0_threshold_20_10_window1_L08_w0_epoch30-detail.jpg}
%\\
%
%\includegraphics[width=2.8cm]{blur10-500-7-input.jpg}
%&\includegraphics[width=2.8cm]{blur10-500-7-input-predict_standard_blur_L2_weight5_window0_threshold_20_10_window2_L08_w0_epoch30.jpg}
%&\includegraphics[width=2.8cm]{blur10-500-7-input-predict_standard_saliency_sharpen_L2_weight5_window0_threshold_20_10_window1_L08_w0_epoch30-detail.jpg}
%\\
%
%\includegraphics[width=2.8cm]{demo11-500-22-input.jpg}
%&\includegraphics[width=2.8cm]{demo11-500-22-input-predict_standard_blur_L2_weight5_window0_threshold_20_10_window2_L08_w0_epoch30.jpg}
%&\includegraphics[width=2.8cm]{demo11-500-22-input-predict_standard_saliency_sharpen_L2_weight5_window0_threshold_20_10_window1_L08_w0_epoch30-detail.jpg}
%\\

\end{tabular}
\end{center}
\vspace{-2mm}
\caption{Content-aware image manipulation with our method. The saliency maps generated by \cite{zhang2017amulet} is only employed in the training stage for the optimization goal; they are not used in the test stage and are presented here for comparison purpose.
Our method can implicitly learn saliency information and produce quality smoothing results accordingly.
%We don't predict saliency map, while which is learned implicitly in deep network and reflected in the predicted smooth image. 
Photo courtesy of Lisa Beattie and Albert Froneman.}
\label{figure:background2}
\vspace{0mm}
\end{figure}

\subsection{Texture Removal}\label{sec:textureremoval}

The texture removal task we consider here aims at removing the fine-scale repetitive patterns from primary image structures. In this task, the smoothing algorithms should be made scale-aware, as the textures to be removed may also have local gradients.

Our method can be easily tailored for this task. To grant a network the ability to distinguish fine-scale textures from primary image structures and smooth them out after training, we can simply set the edge responses of the texture points on the guidance image $E(I)$ to be zero. This way, the corresponding edge responses on the guidance map of the output image $E(T)$ will always be larger. Thus with slight modification on the constraint of Equation \ref{eq:p}, a $L_2$ smoothness regularizer can be easily enforced on the texture regions, such that the network will learn to diminish them. The way to obtain the texture structure is elaborated in the supplemental material.

Figure~\ref{figure:texture} shows two examples that contain different types of texture patterns. We compare our results against state-of-the-art methods of \cite{RTV,RGF,cho2014bilateral,ham2015robust,zhang2015segment} that address the texture removal problem. It can be seen that both \cite{RGF} and \cite{cho2014bilateral} tends to blur some large-scale major structures, while the method of \cite{ham2015robust} produces some noisy structure boundaries. Compared to these methods, superior results are obtained by our method.

Since this task aims at diminishing textures that are very possibly locally salient, we enlarge the $p^{large}$ weight ($\alpha = 20$) and limit the smooth region ($h = 5$).

\subsection{Content-Aware Image Manipulation}\label{sec:contentaware}
Different from traditional methods, our proposed algorithm enables us to achieve content-aware image processing, \emph{i.e.}, smoothing a particular category of objects in the image.

In this section, we use the image saliency cue to demonstrate content-aware image manipulations by our method. For example, the proposed objective function can be slightly modified to achieve background smoothing goal, which is smoothing out the background regions for highlighting the foreground (\emph{i.e.}, the salient objects). To this end, we mask out the edge responses of the background regions in the guidance image $E(I)$ via the binary saliency masks obtained by recent salient object detection algorithm \cite{zhang2017amulet}. By feeding the modified guidance image $E(I)$ to the proposed objective function, the $L_2$ norm regularizer can be applied on the background regions during training. Afterwards, we can even set the smoothing weights of foreground regions to relatively small values or even zero to keep the foreground unmodified. Figure~\ref{figure:background2} presents some example results from our method, from which we can see that our trained network is capable of implementing content-aware image smoothing very well.

Alternatively, our algorithm are also able to smooth out foreground regions via a similar strategy, such that a foreground enhancement effect can be achieved via the approach described in Section~\ref{detail}. Figure~\ref{figure:background2} demonstrates very visually-pleasing exaggeration effect for the foreground objects via our approach.

Note in this application, the smoothness effects and saliency information are jointly learned within our network, while the latter information is reflected in the predicted smooth image. All these results are obtained solely by our trained network without any pre- or post- processing. We set $h$ in Equation \ref{eq:Ef2} as 5 to limit the smoothness only within either the foreground or background region. 

\setlength{\tabcolsep}{1pt}
\begin{figure}[t]
\begin{center}
\begin{tabular}{cc}
\includegraphics[width=4.2cm]{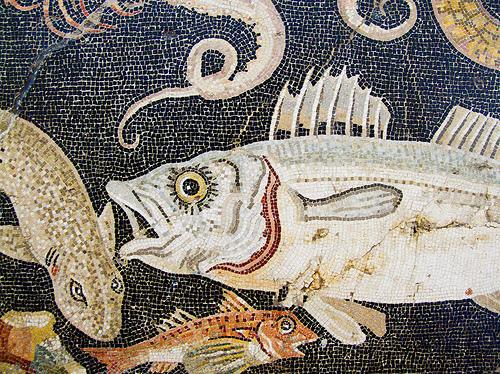}
&\includegraphics[width=4.2cm]{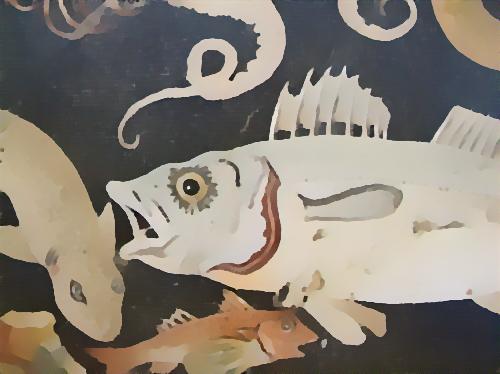}
\end{tabular}
\end{center}
\vspace{-2mm}
\caption{A partial failure case of texture-removal. Our algorithm doesn't succeed in extracting some detailed textures perfectly such as the eyes of the two smaller fishes. Photo courtesy of Flickr user Chris Beckett.}
\label{figure:failure}
\end{figure}

\section{Conclusion}
In this paper, we have presented an unsupervised learning approach for the task of image smoothing. We introduced a novel image smoothing objective function built upon the mixture of a spatially-adaptive $L_p$ flattening criterion and an edge-preserving regularizer. These criteria not only lead to state-of-the-art smoothing effects as demonstrated in our experiments, but also grant our method the flexibility to obtain different smoothing effects within a single framework. We have also shown that training a deep neural network on a large corpus of raw images without ground truth labels can adequately solve the underlying minimization problem and generate impressive results. Moreover, the end-to-end mapping from a single input image to its corresponding smoothed counterpart by the neural network can be computed efficiently on both GPU and CPU, and the experiments have shown that our method runs orders of magnitude faster than traditional methods. We foresee a wide range of applications that can benefit from our new pipeline.

\vspace{0pt}
\subsection{Limitations and Future work}

Our algorithm relies on some additional information to optimize the objective function during training, such as the detected structures or textures. Currently we employ some simple heuristic methods to detect the structures, and imperfect detection can influence the smoothing results. Figure~\ref{figure:failure} shows an example where our algorithm fails to extract some detailed textures perfectly. This issue can mitigated by introducing moderate effort of human interaction for refining the structure maps of the training data, or by synthesizing training images with separate textures and clear images similar to \cite{kaiyue}. Developing more advanced detection algorithms is also one of our future works.

%However, this shortcoming can be eased with moderate effort of human interaction by removing the imperfect extracted texture map from the training data, or synthesizing training images with separate textures and clear images similar to \cite{kaiyue}.

%extract some detailed textures on fish eyes perfectly. Since structure/edge extraction is not the focus of this paper, developing more advanced methods for it is our future work. 

Due to the adaptively changed and spatially variant $L_p$ flattening term and extra input information required for optimization, the optimization is complex and very challenging for traditional numerical solvers. To our knowledge, this is the first attempt of treating deep network as a numerical solver in the image smoothing field. In the future, we also would like to explore more complex and different tasks, such as multi-image or video processing.

%Moreover, it usually demands for iterative optimization over a non-convex objective function, which is also insufficient for some practical real-time applications. The above shortcomings provide an opportunity for the deep neural network to serve as a solver instead of the common approximator to tackle the optimization problems.

\begin{acks}
The authors would also like to thank the anonymous reviewers for their valuable comments and helpful suggestions. This work is supported by the \grantsponsor{GS501100001809}{National 973 Program}{} under Grant No. 2015CB352501, and \grantsponsor{GS501100001810}{NSFC-ISF}{} under Grant No. 61561146397.
\end{acks}

%\section*{Acknowledgements}
%\begin{acks}
%We thank the anonymous reviewers for their helpful comments. This work was supported by National 973 Program (2015CB352501) and NSFC-ISF (61561146397).
%\end{acks}

\bibliographystyle{ACM-Reference-Format}
\bibliography{template}

\end{document}